\documentclass[review]{fcs}
\usepackage{float}
\usepackage{xcolor}
\usepackage{bm}
\usepackage{multirow}
\usepackage{makecell}
\definecolor{White}{rgb}{1.,0.,1.}
\definecolor{first}{rgb}{.8,.0,.0}
\definecolor{second}{rgb}{.0,.6,.0}
\definecolor{third}{rgb}{.0,.0,.8}
\newcolumntype{g}{>{\columncolor{White}}c}

\definecolor{car}{rgb}{0.39215686, 0.58823529, 0.96078431}
\definecolor{bicycle}{rgb}{0.39215686, 0.90196078, 0.96078431}
\definecolor{motorcycle}{rgb}{0.11764706, 0.23529412, 0.58823529}
\definecolor{truck}{rgb}{0.31372549, 0.11764706, 0.70588235}
\definecolor{othervehicle}{rgb}{0.39215686, 0.31372549, 0.98039216}
\definecolor{person}{rgb}{1.        , 0.11764706, 0.11764706}
\definecolor{bicyclist}{rgb}{1.        , 0.15686275, 0.78431373}
\definecolor{motorcyclist}{rgb}{0.58823529, 0.11764706, 0.35294118}
\definecolor{road}{rgb}{1.        , 0.        , 1.        }
\definecolor{parking}{rgb}{1.        , 0.58823529, 1.        }
\definecolor{sidewalk}{rgb}{0.29411765, 0.        , 0.29411765}
\definecolor{otherground}{rgb}{0.68627451, 0.        , 0.29411765}
\definecolor{building}{rgb}{1.        , 0.78431373, 0.        }
\definecolor{fence}{rgb}{1.        , 0.47058824, 0.19607843}
\definecolor{vegetation}{rgb}{0.        , 0.68627451, 0.        }
\definecolor{trunk}{rgb}{0.52941176, 0.23529412, 0.        }
\definecolor{terrain}{rgb}{0.58823529, 0.94117647, 0.31372549}
\definecolor{pole}{rgb}{1.        , 0.94117647, 0.58823529}
\definecolor{trafficsign}{rgb}{1.        , 0.        , 0.    }

\definecolor{nbarrier}{RGB}{255, 120, 50}
\definecolor{nbicycle}{RGB}{255, 192, 203}
\definecolor{nbus}{RGB}{255, 255, 0}
\definecolor{ncar}{RGB}{0, 150, 245}
\definecolor{nconstruct}{RGB}{0, 255, 255}
\definecolor{nmotor}{RGB}{200, 180, 0}
\definecolor{npedestrian}{RGB}{255, 0, 255}
\definecolor{ntraffic}{RGB}{255, 240, 150}
\definecolor{ntrailer}{RGB}{135, 60, 0}
\definecolor{ntruck}{RGB}{255, 0, 0}
\definecolor{ndriveable}{RGB}{213, 213, 213}
\definecolor{nother}{RGB}{139, 137, 137}
\definecolor{nsidewalk}{RGB}{75, 0, 75}
\definecolor{nterrain}{RGB}{150, 240, 80}
\definecolor{nmanmade}{RGB}{160, 32, 240}
\definecolor{nvegetation}{RGB}{0, 175, 0}
\definecolor{nothers}{RGB}{0, 0, 0}

\makeatletter
\newcommand{\car@semkitfreq}{3.92}
\newcommand{\bicycle@semkitfreq}{0.03}
\newcommand{\motorcycle@semkitfreq}{0.03}
\newcommand{\truck@semkitfreq}{0.16}
\newcommand{\othervehicle@semkitfreq}{0.20}
\newcommand{\person@semkitfreq}{0.07}
\newcommand{\bicyclist@semkitfreq}{0.07}
\newcommand{\motorcyclist@semkitfreq}{0.05}
\newcommand{\road@semkitfreq}{15.30}  %
\newcommand{\parking@semkitfreq}{1.12}
\newcommand{\sidewalk@semkitfreq}{11.13}  %
\newcommand{\otherground@semkitfreq}{0.56}
\newcommand{\building@semkitfreq}{14.1}  %
\newcommand{\fence@semkitfreq}{3.90}
\newcommand{\vegetation@semkitfreq}{39.3}  %
\newcommand{\trunk@semkitfreq}{0.51}
\newcommand{\terrain@semkitfreq}{9.17} %
\newcommand{\pole@semkitfreq}{0.29}
\newcommand{\trafficsign@semkitfreq}{0.08}
\newcommand{\semkitfreq}[1]{{\csname #1@semkitfreq\endcsname}}

\title{Vision-based 3D occupancy prediction in autonomous driving: a review and outlook}
\author*{Yanan ZHANG, Jinqing ZHANG, Zengran WANG, Junhao XU, Di HUANG}
\address{Laboratory of Intelligent Recognition and Image Processing, School of Computer Science and Engineering, Beihang University, Beijing 100191, China}
\fcssetup{
  received       = {month dd, yyyy},
  accepted       = {month dd, yyyy},
  corr-email     = {dhuang@buaa.edu.cn},
}
\begin{abstract}
In recent years, autonomous driving has garnered escalating attention for its potential to relieve drivers' burdens and improve driving safety. Vision-based 3D occupancy prediction, which predicts the spatial occupancy status and semantics of 3D voxel grids around the autonomous vehicle from image inputs, is an emerging perception task suitable for cost-effective perception system of autonomous driving. Although numerous studies have demonstrated the greater advantages of 3D occupancy prediction over object-centric perception tasks, there is still a lack of a dedicated review focusing on this rapidly developing field. In this paper, we first introduce the background of vision-based 3D occupancy prediction and discuss the challenges in this task. Secondly, we conduct a comprehensive survey of the progress in vision-based 3D occupancy prediction from three aspects: feature enhancement, deployment friendliness and label efficiency, and provide an in-depth analysis of the potentials and challenges of each category of methods. Finally, we present a summary of prevailing research trends and propose some inspiring future outlooks. To provide a valuable reference for researchers, a regularly updated collection of related papers, datasets, and codes is organized at \url{https://github.com/zya3d/Awesome-3D-Occupancy-Prediction}.
\end{abstract}

\begin{document}
\setlength{\baselineskip}{14pt}
%\sloppy 

\section{Introduction}

In recent years, with the development of deep learning and sensor technology, autonomous driving has made significant progress, promising to effectively reduce traffic accidents, improve transportation efficiency and enhance people's travel experience. Autonomous driving systems typically comprise three crucial modules: environment perception, behavior decision and motion control. The environment perception module analyzes data captured by onboard sensors to perceive and understand the surrounding environment, serving as a prerequisite for behavior decision and motion control,  and holding a vital role in autonomous driving.

3D object detection, as a fundamental component of autonomous driving perception systems, has drawn widespread attention from researchers, resulting in three branches based on diverse sensor data modalities: image-based methods~\cite{chen2016monocular,chabot2017deep,wang2019pseudo,li2019stereo,chen2020dsgn,ma20233d}, LiDAR-based methods~\cite{yan2018second,yin2021center,deng2021voxel,shi2019pointrcnn,zhang2021pc,zhou2023octr}, and multi-modal methods~\cite{vora2020pointpainting,pang2020clocs,chen2017multi,huang2020epnet,zhang2022cat,wang2023multi}. While LiDAR-based and multi-modal methods can achieve relatively strong detection performance, image-based methods are favored for practical deployment due to their lower economic costs and better real-time capabilities. Recently, the emergence of the Bird’s-Eye-View (BEV) representation paradigm~\cite{huang2021bevdet,li2023bevdepth,li2022bevformer,zhang2023sa,ma2022vision} has improved the accuracy of vision-centric 3D object detection, further narrowing the performance gap.

However, vision-centric 3D object detection based on BEV representation still faces inherent limitations in complex open scenarios: (1) limited representation. The 3D bounding box only estimates the maximum possible boundary of the foreground objects, and this coarse-grained representation cannot depict the fine-grained internal geometric shape and model the background areas. Additionally, Bird's Eye View (BEV) representation essentially provides a top-down projection of 3D space, inevitably resulting in information loss in terms of height. (2) limited detection. In complex scenes, factors like occlusion, lighting variations, and noise are commonly present, and BEV-based detections are particularly susceptible to interference, resulting in the loss of entire objects despite some features remaining available. Furthermore, in open scenarios, long-tail objects with undefined shapes or appearances, such as deformable obstacles like excavators and trailers, as well as obstacles of unknown categories like vegetation, gravel, and garbage, pose challenges for encompassing all corner cases comprehensively.

Vision-based 3D occupancy prediction, which predicts the spatial occupancy status and semantic categories of 3D voxel grids around the autonomous vehicle from image inputs, is a promising solution for providing fine-grained representation and robust detection for undefined long-tail obstacles in 3D space. Occupancy representation originated from the field of robotics, where 3D space is divided into voxel units for binary prediction of whether voxels are occupied by objects, enabling effective collision avoidance. Mescheder et al.~\cite{mescheder2019occupancy} propose the Occupancy Network, which implicitly represents 3D surfaces via the continuous decision boundary of a deep neural network classifier, utilizing occupancy grid mapping to determine whether grids are occupied. Peng et al.~\cite{peng2020convolutional} further introduce Convolutional Occupancy Networks, a novel shape representation which combines the expressiveness of convolutional neural networks with the advantages of implicit representations. At the Tesla AI Day 2022, Tesla introduces Occupancy Network to autonomous driving, sparking a research wave in vision-based 3D occupancy prediction. 

In this paper, we comprehensively review the vision-based 3D occupancy prediction methods for autonomous driving applications and systematically categorize existing methods into three groups: feature enhancement methods, deployment-friendly methods and label-efficient methods. For feature enhancement, we further summarize the feature representation methods based on Bird’s Eye View (BEV), Tri-Perspective View (TPV) and 3D Voxel. From computational friendliness, we conduct an in-depth analysis from two viewpoints: perspective decomposition and coarse-to-fine paradigm. As for label efficiency, we methodically categorize relevant research into annotation-free methods and LiDAR-free methods. Finally, we provide some inspiring future outlooks for 3D occupancy prediction from the perspectives of data, methodology, and task.

The major contributions of this paper are as follows:

\begin{itemize}
\item To the best of our knowledge, this paper is the first comprehensive review tailored to the vision-based 3D occupancy prediction methods for autonomous driving.
\item This paper structurally summarizes vision-based 3D occupancy prediction methods from three views: feature-enhanced, computation-friendly and label-efficient approaches, and provides in-depth analysis and comparison for different categories of methods.
\item The paper proposes some inspiring future outlooks for vision-based 3D occupancy prediction and provides a regularly updated github repository to collect the related papers, datasets, and codes.
\end{itemize}

\begin{figure*}[!ht]
\centering
\includegraphics[width=1.0\linewidth]{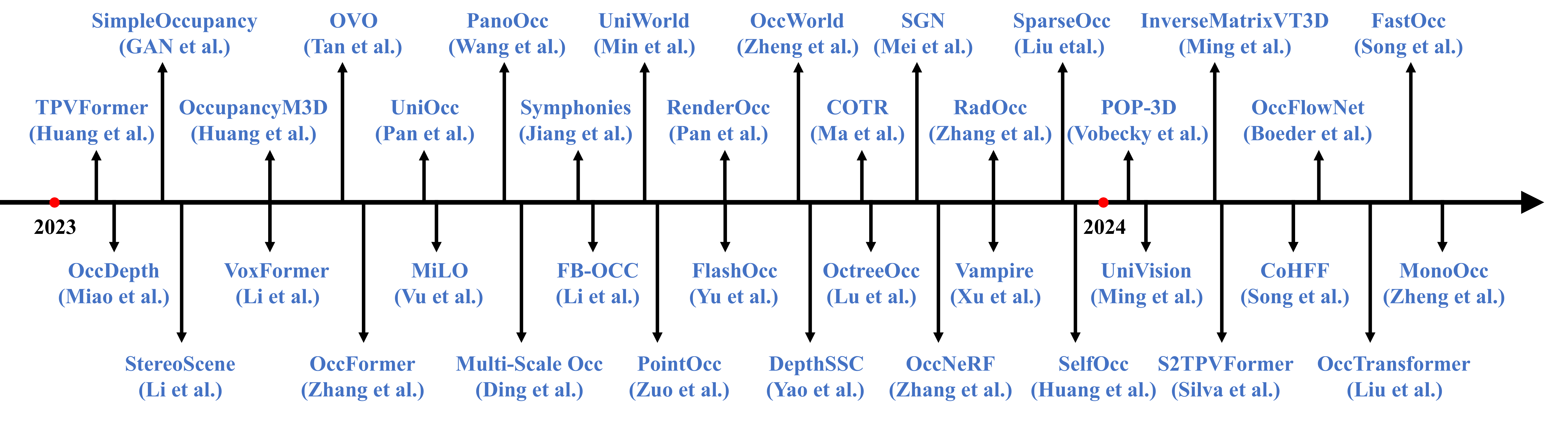}
\caption{Chronological overview of vision-based 3D occupancy prediction methods.}
\label{timeline}
\end{figure*}

\begin{figure*}[!ht]
\centering
\includegraphics[width=0.99\linewidth]{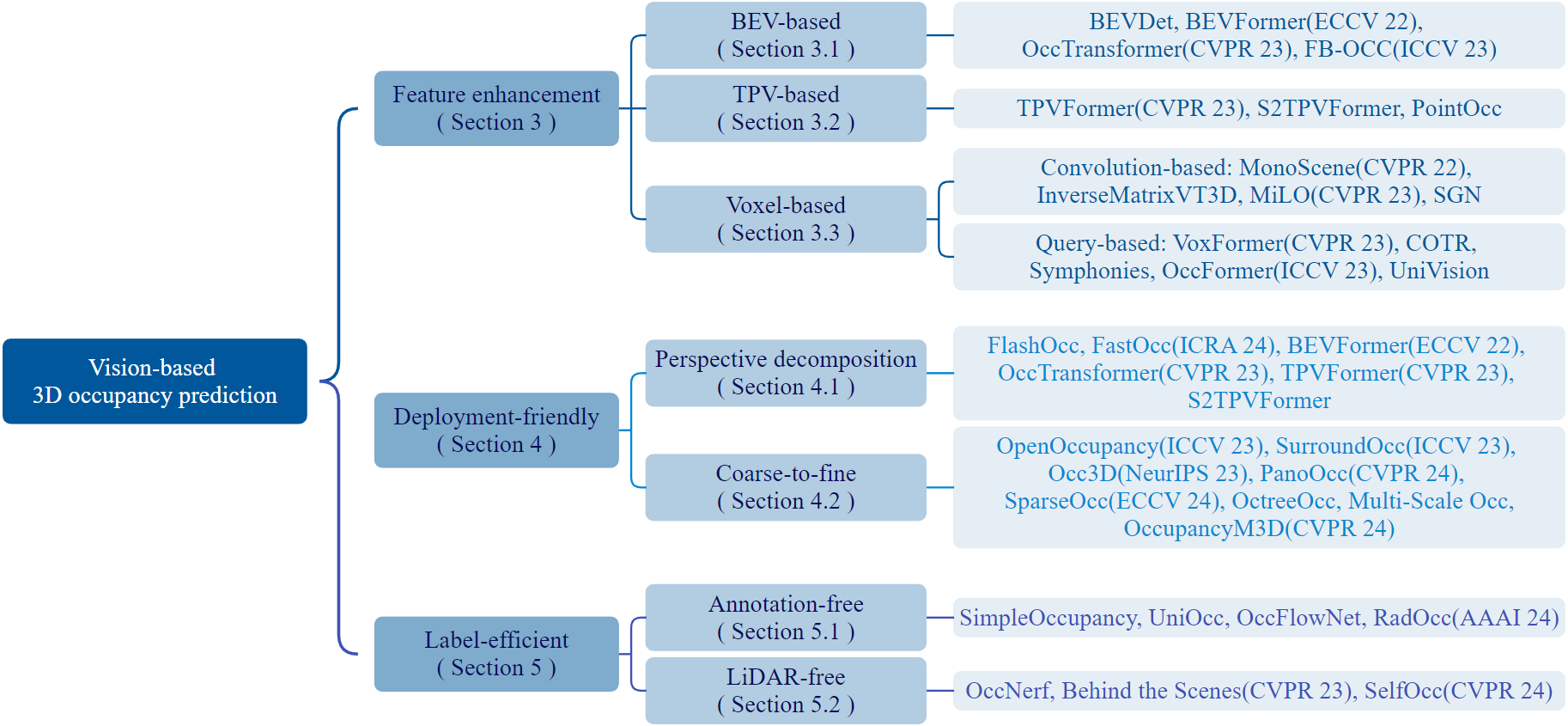}
\caption{Hierarchically-structured taxonomy of vision-based 3D occupancy prediction for autonomous driving.}
\label{overall}
\end{figure*}

The structure of this paper is organized as follows. First, we introduce the problem definition, datasets, evaluation metrics and key challenges of vision-based 3D occupancy prediction in Section~\ref{BG}. Then, we review and analyze the vision-based 3D occupancy prediction methods based on feature enhancement (Section~\ref{FE}), computational friendliness (Section~\ref{CF}), and label efficiency (Section~\ref{LE}). Finally, we investigate the research trends, and prospect the future directions of vision-based 3D occupancy prediction in Section~\ref{FO}. A chronological overview of vision-based 3D occupancy prediction methods is shown in Figure~\ref{timeline} and the corresponding hierarchically-structured taxonomy is shown in Figure~\ref{overall}. 

\section{Background}
\label{BG}

This section elaborates on the task definition, ground-truth generation, common datasets, evaluation metrics and key challenges in the domain of vision-based 3D occupancy prediction.

\subsection{The definition of vision-based 3D occupancy prediction}

Given images from multi-view cameras $\bm{I}=\{ \bm{I}_{1}, \bm{I}_{2}, ... , \bm{I}_{n} \}$, where $n$ denotes the number of camera views, the model $\mathcal{F}_{\theta}$ needs to jointly predict the occupancy state of each voxel in the scene in range $[H_l, W_l, Z_l, H_r, W_r, Z_r]$. That is, classify the voxel into two states: empty or occupied $\bm{O_o}\in \{ 0, 1\}^{H\times W\times Z}$, where $H, W, Z$ denote the number of voxels in $x, y, z$ axis and each voxel has the shape of $[\frac{W_l - W_r}{W}, \frac{H_l - H_r}{H}, \frac{Z_l - Z_r}{Z}]$. If a voxel is predicted to be occupied, its semantic class also needs to be predicted among predefined classes $\bm{O_{c}}\in \{1, 2, ... , C\}^{N}$, where $C$ denote the number of predefined classes and $N$ denote the number of occupied voxels. The two outputs are combined as $\mathcal{F}_{\theta}(\bm{I})=\bm{O_o}\in \{0, 1, 2, ... , C\}^{H\times W\times Z}$. For each input frame, a visibility mask $\bm{M}\in \{ 0, 1\}^{H\times W\times Z}$ is also generally provided, and invisible voxels will be ignored during evaluation. Training a vision-based 3D occupancy prediction model can be described as: minimize a total loss $\mathcal{L}$, which is generally a classification loss, across a set of multiple-view cameras images and its corresponding annotated occupancy label $\bm{y}\in \{0, 1, 2, ... , C\}^{H\times W\times Z}$ from the dataset $\mathcal{D}$:
\begin{equation}
\min_{\theta} \mathcal{L}( \mathcal{F}_{\theta}(\bm{I})\otimes \bm{M}, \bm{y}\otimes \bm{M} ), \ \ where\ \ (\bm{I}, \bm{y}, \bm{M})\in \mathcal{D}.
\end{equation}

\subsection{Ground truth generation}

Generating ground truth labels is a challenge for 3D occupancy prediction. Although many 3D perception datasets, such as nuScenes\cite{caesar2020nuscenes} and Waymo\cite{sun2020scalability}, provide LiDAR points segmentation labels, these labels are sparse and difficult to supervise dense 3D occupancy prediction tasks. Wei et al. \cite{wei2023surroundocc} have demonstrated the importance of using dense occupancy as ground truth. Some recent researches\cite{wei2023surroundocc, wang2023openoccupancy, tian2023occ3d, tong2023scene} have focused on generating dense 3D occupancy annotations using sparse LiDAR points segmentation annotations, providing some useful datasets and benchmarks for 3D occupancy prediction tasks.

\begin{figure}[!ht]
\centering
\includegraphics[width=.99\linewidth]{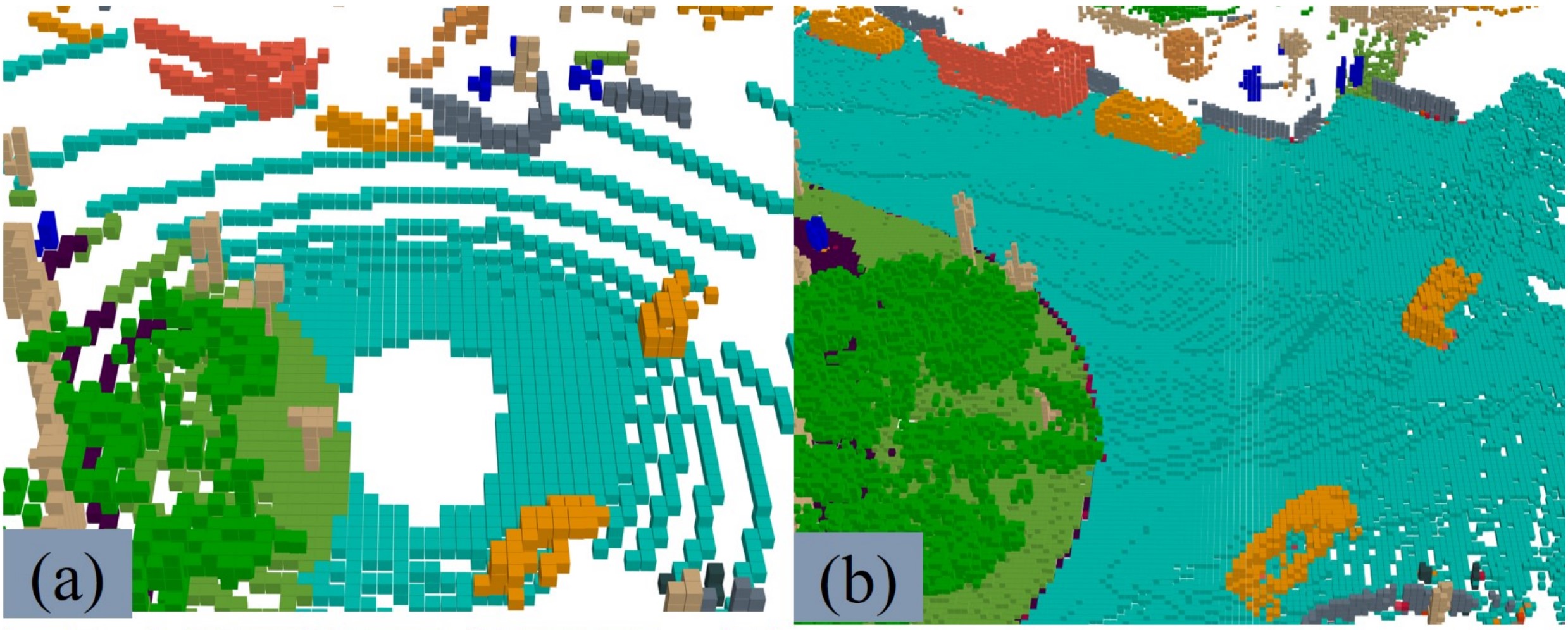}
\caption{Visual comparison on 3D occupancy annotations\cite{tong2023scene}. (a) sparse occupancy; (b) dense occupancy.}
\label{fig:2}
\end{figure}

The ground truth labels in the 3D occupancy prediction task present whether each voxel in the 3D space is occupied and the semantic label of the occupied voxel. Due to the large number of voxels in 3D space, it is very difficult to manually label each voxel. A common practice is to voxelize the ground truth of the existing 3D point cloud segmentation task and then generate the ground truth of 3D occupancy prediction by voting according to the semantic labels of the points in the voxel. However, the ground truth generated by this way is sparse. As shown in Figure \ref{fig:2}, there are still many occupied voxels in places such as the roads that are not marked as occupied. Supervising models with such sparse ground truth will lead to a decrease in model performance. Therefore, some works investigate how to automatically or semi-automatically generate high-quality dense 3D occupancy annotations.

As shown in Figure \ref{fig:3}, generating dense 3D occupancy annotations typically involves the following four steps:

\begin{figure*}[!ht]
\centering
\includegraphics[width=0.99\linewidth]{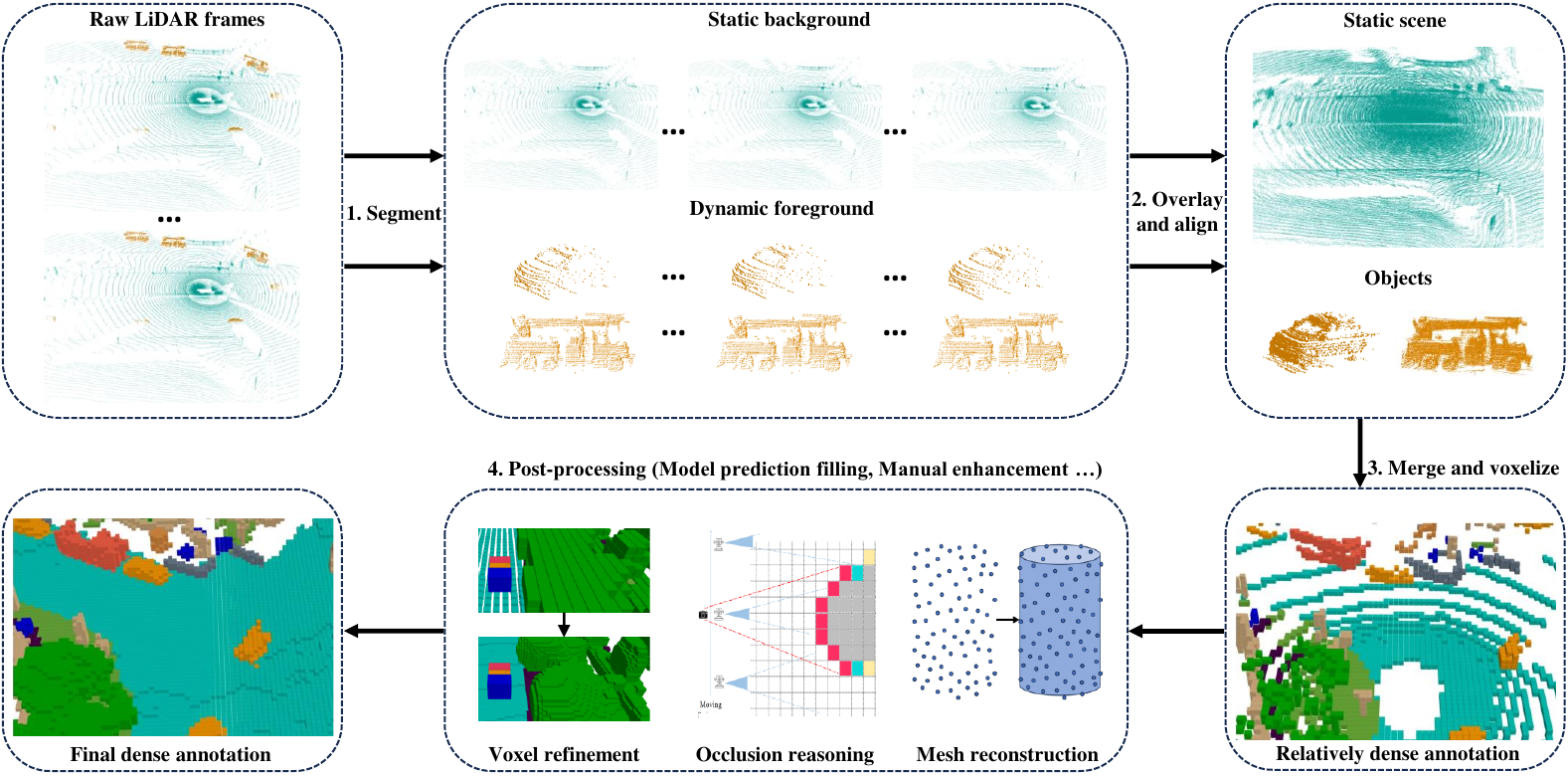}
\caption{The pipeline for generating dense 3D occupancy annotations.}
\label{fig:3}
\end{figure*}

\begin{itemize}
\item[1)] Take consecutive raw LiDAR frames and segment the LiDAR points into the static background and the dynamic foreground.
\item[2)] Overlay continuous LiDAR frames on the static background and perform motion compensation based on positioning information to align multi-frame point cloud for obtaining a more dense point cloud. Superimpose continuous LiDAR frames on the dynamic foreground, and the point cloud of the dynamic foreground is aligned according to the target frame and target id to make it more dense. Note that although the point cloud is relatively denser, there are still some gaps after voxelization, which need further processing.
\item[3)] Merge the foreground and background point clouds, then voxelize them and use a voting mechanism to determine the semantics of the voxels, resulting in a relatively dense voxel annotation.
\item[4)] Refining the voxels obtained in the previous step through post-processing to achieve denser and finer annotations, serving as the ground truth.
\end{itemize}

Different works are essentially identical in steps 1) and 2), with the main differences focusing on steps 3) and 4), which are explained in detail below.

Wang et al. \cite{wang2023openoccupancy} propose the APP (Augmenting And Purifying) pipeline for generating dense 3D occupancy annotations. In the post-processing step, inspired by self-training, they use the relatively dense annotations obtained in step 3 as the initial labels, and use the initial labels to supervise the training to obtain a coarse model. They then supplement the output of the model with the voxels marked empty in the initial label for augmentation. Finally, they design a labeling software and use manual enhancement to purify the labels, which took about 4,000 human hours. This semi-automated approach results in high quality annotations, but it also brings a lot of human cost, so subsequent research rarely use this human purification.

Tian et al. \cite{tian2023occ3d} further propose a semi-automatic label generation pipeline. After voxelization in step 3, they utilize K-Nearest Neighbor voting to determine the semantic labels of voxels. In the post-processing step, mesh reconstruction methods such as VDBFusion\cite{vizzo2022vdbfusion} is used to obtain denser voxel annotations. Additionally, the corresponding LiDAR visibility and camera visibility masks are generated for 3D annotations according to the geometric principle. Finally, to address the issue of object 3D shape enlargement caused by lidar noise and localization errors, they have introduced an image-guided voxel refinement method to eliminate incorrectly occupied voxels under the guidance of image semantic segmentation masks. This method significantly enhances the shape of object boundaries and improves the quality of annotation. Note that they also propose a method to quantitatively evaluate the quality of generated labels by checking the semantic consistency between 2D pixels and 3D voxels, and conduct ablation experiments to demonstrate the effectiveness of each step in their pipeline.

Wei et al. \cite{wei2023surroundocc} employ the nearest neighbor algorithm to assign semantic labels to each voxel, and use Poisson reconstruction to fill the holes during post-processing in order to obtain dense 3D occupancy labels. Tong et al. \cite{tong2023scene} determine the semantics of voxels by majority voting of point semantic labels within voxels, and utilize unlabeled intermediate frames and scene completion to enhance voxel density for obtaining dense annotations during post-processing. It is worth noting that they also annotate the flow velocity of the foreground voxels based on the velocity of the 3D boxes for more downstream tasks such as motion planning.

\subsection{Datasets}
In this subsection, we introduce some open source, large-scale datasets commonly used for 3D occupancy prediction, and a comparison between them is given in Table~\ref{table:table1}.

\begin{table*}[]
\caption{Comparison of datasets for 3D occupancy prediction. Mult-view=$\checkmark$ represents datasets that use mult-view inputs. $C$, $D$, $L$ denote camera, depth and LiDAR. GO is short for General Object.}
\centering
\begin{tabular}{ccccccccc}
\hline  
\textbf{} & \textbf{Year} & \textbf{Type}
& \textbf{Mult-view}
& \textbf{Input} & \textbf{\#Classes} 
& \textbf{\#Sequences} & \textbf{\#Frames}
& \textbf{Voxel Size}\\ 
\hline  
NYUv2\cite{silberman2012indoor} & 2012 & Indoor & $\times$ & C\&D &
13 & 1,449 & 1,449 & 240$\times$240$\times$144\\
SemanticKITTI\cite{behley2019semantickitti} & 2019 & Outdoor & $\times$ & C\&L &
28 & 22 & 43,552 & 256$\times$256$\times$32\\
Occ3D-nuScenes\cite{tian2023occ3d} & 2023 & Outdoor & $\checkmark$ & C &
16+GO & 850 & 40,000 & 200$\times$200$\times$16\\
Occ3D-Waymo\cite{tian2023occ3d} & 2023 & Outdoor & $\checkmark$ & C &
14+GO & 1,000 & 200,000 & 200$\times$200$\times$32\\
nuScenes-Occupancy\cite{wang2023openoccupancy} & 2023 & Outdoor & $\checkmark$ & C\&L &
17 & 850 & 200,000 & 512$\times$512$\times$40\\
OpenOcc\cite{tong2023scene} & 2023 & Outdoor & $\checkmark$ & C\&L &
16 & 850 & 34,149 & 200$\times$200$\times$16\\
\hline
\end{tabular} 
\label{table:table1}
\end{table*}

NUYv2\cite{silberman2012indoor} dataset consists of video sequences from various indoor scenes, captured by both the RGB and Depth cameras of the Microsoft Kinect. It contains 1449 densely labeled pairs of aligned RGB and depth images as well as 407,024 unlabeled frames from 3 cities. While primarily intended for indoor use and not ideal for autonomous driving scenarios, some studies have utilized this dataset for 3D occupancy prediction.

SemanticKITTI\cite{behley2019semantickitti} is a widely used dataset for 3D occupancy prediction, comprising 22 sequences and over 43,000 frames from the KITTI\cite{geiger2012we} dataset. It creates dense 3D occupancy annotations by overlaying future frames, segmenting voxels, and assigning labels through point voting. Additionally, it check for every pose of the car which voxels are visible to the sensor by tracing a ray and ignore the invisible voxels during training and evaluation. However, as it is based on the KITTI dataset, it only utilizes images from the front-facing camera as input, while subsequent datasets often employ multi-view images. As shown in Table~\ref{table:table2}, we collect the evaluation results of existing methods on SemanticKITTI dataset. 

NuScenes-occupancy\cite{wang2023openoccupancy} is a 3D Occupancy prediction dataset constructed based on nuScenes\cite{caesar2020nuscenes}, an large-scale autonomous driving dataset for outdoor environments. It contains 850 sequences, 200,000 frames, and 17 semantic categories. The dataset initially produces rough 3D occupancy labels using an Augmenting And Purifying (AAP) pipeline, and then refines the labels through manual augmentation. Additionally, it introduces OpenOccupancy, the first benchmark for surrounding semantic occupancy perception, to assess advanced 3D occupancy prediction methods.

Subsequently, Tian et al. \cite{tian2023occ3d} further built the Occ3D-nuScenes and Occ3D-Waymo datasets for 3D occupancy prediction based on nuScenes and Waymo\cite{sun2020scalability} autonomous driving datasets. They introduce a semi-automatic label generation pipeline that utilizes existing labeled 3D perception datasets and identifies voxel types based on their visibility. Additionally, they establish the Occ3d benchmark for large-scale 3D occupancy prediction to enhance the evaluation and comparison of different methods. As shown in Table~\ref{table:table2}, we collect the evaluation results of existing methods on Occ3D-nuScenes dataset.

Additionally, similar to Occ3D-nuScenes and Nuscenes-Occupancy, OpenOcc\cite{tong2023scene} is also a dataset built for 3D Occupancy prediction based on the nuScenes dataset. It contains 850 sequences, 34,149 frames, and 16 classes. Note that this dataset provides the additional flow annotation of eight foreground objects, which is helpful for the downstream task such as motion planning.

\subsection{Evaluation metrics}
Similar to 2D pixel-level semantic segmentation tasks, 3D occupancy prediction commonly utilizes Mean Intersection over Union ($mIoU$) as a metric to assess the semantic segmentation performance of models:
\begin{equation}
mIoU=\frac{1}{C}\sum_{c=1}^{C}{\frac{TP_{c}}{TP_{c}+FP_{c}+FN_{c}}}
\end{equation}
where $TP_{c}$, $FP_{c}$, and $FN_{c}$ denote the true positives, false positives and false negatives for class $c$, respectively. For vision-centric task, datasets often provide a camera visibility mask to compute the mIoU on regions that are visible in images.

Intersection over union ($IoU$) is also used to evaluate the scene completion quality regardless of semantic labels:
\begin{equation}
IoU=\frac{TP_{o}}{TP_{o}+FP_{o}+FN_{o}}
\end{equation}
where $TP_{o}$, $FP_{o}$, and $FN_{o}$ denote the true positives, false positives and false negatives for occupancy voxels.

Note that $IoU$ and $mIoU$ are two different metrics to evaluate the performance of scene completion and semantic segmentation. There is a strong interaction between $IoU$ and $mIoU$, e.g., a high $mIoU$ can be achieved by naively decreasing the $IoU$\cite{li2023voxformer}, so a good model should perform well on both metrics.

\begin{table*}[!ht]
\caption{3D occupancy prediction performance on SemanticKITTI and Occ3D-nuScenes datasets. The best one is highlighted in \textbf{bold}. Note that some experimental settings may not be exactly the same, and the results are just collected here for reference.}
\centering

%\scalebox{0.9}{
\begin{tabular}{c|c|c|c|c|c|c|c|c}
\hline
\multicolumn{1}{c|}{\multirow{2}{*}{Category}} & \multirow{2}{*}{Method}               & \multirow{2}{*}{Published in} & \multicolumn{4}{c|}{SemanticKITTI}& \multicolumn{2}{c}{Occ3D-nuScenes}\\ \cline{4-9}
\multicolumn{1}{c|}{}                          &                                       &                       & IoU & IoU rank & mIoU & mIoU rank & mIoU          & mIoU rank          \\ \hline
\multirow{14}{*}{Feature enhancement methods}           & BEVDet\cite{huang2021bevdet}          & arXiv 2021            &--   &--        &--    &--             & 19.38         & 19/22             \\ %\cline{2-9}
& JS3CNet\cite{yan2021sparse}           & AAAI 2021             &34.00& 11/14    & 8.97 & 14/17         & --            & --                \\ %\cline{2-9}
                                                & MonoScene\cite{cao2022monoscene}      & CVPR 2022             &34.16& 10/14    &11.08 & 12/17         & 6.06          & 22/22             \\ %\cline{2-9}
                                                & BEVFormer\cite{li2022bevformer}       & ECCV 2022            &--   &--        &--    &--             & 26.88         & 16/22             \\ %\cline{2-9}
                                                & BEVStereo\cite{li2023bevstereo}       & AAAI 2023             &--   &--        &--    &--             & 42.02         & 10/22             \\ %\cline{2-9}
                                                & TPVFormer\cite{huang2023tri}          & CVPR 2023             &34.25& 9/14     &11.26 & 11/17         & 28.34         & 15/22             \\ %\cline{2-9}
                                                & VoxFormer\cite{li2023voxformer}       & CVPR 2023             &42.95& 5/14     &12.20 &  9/17         & --            & --                \\ %\cline{2-9}
                                                & OccFormer\cite{zhang2023occformer}    & ICCV 2023             &34.53& 8/14     &12.32 &  8/17         & 21.93         & 18/22             \\ %\cline{2-9}
                                                & Symphonize\cite{jiang2023symphonize}  & arXiv 2023            &42.19& 6/14     &15.04 &  4/17         & --            & --                \\ %\cline{2-9}
                                                & SGN\cite{mei2023camera}               & arXiv 2023            &\textbf{45.42}& 1/14     &\textbf{15.76} &  1/17         & --            & --                \\ %\cline{2-9}
                                                & MiLO\cite{myeongjin2023milo}          & arXiv 2023            &--   &--        &--    &--             & 52.45         &  2/22             \\ %\cline{2-9}
                                                & FB-OCC\cite{li2023fb}                 & ICCV 2023            &--   &--        &--    &--             & \textbf{52.79}         &  1/22             \\ %\cline{2-9}
                                                & StereoScene\cite{li2023stereoscene}   & IJCAI 2024            &43.34& 4/14     & 15.36&  3/17         & --            & --                \\ %\cline{2-9}
                                                & OccTransformer\cite{liu2024occtransformer} & arXiv 2024       &--   &--        &--    &--             & 49.23         &  6/22             \\ \hline
                                                
\multirow{13}{*}{Deployment-friendly methods} 
&3DSketch\cite{chen20203d}             & CVPR 2020             &26.85& 13/14    & 6.23 & 17/17         & --            & -- 
\\ %\cline{2-9}
& AICNet\cite{li2020anisotropic}        & CVPR 2020             &23.93& 14/14    & 7.09 & 15/17         & --            & --                \\ %\cline{2-9}
& LMSCNet\cite{roldao2020lmscnet}       & 3DV 2020            &31.38& 12/14    & 7.07 & 16/17         & --            & --                \\ %\cline{2-9}
& CTFOcc\cite{tian2023occ3d}            & NIPS 2023             &--   &--        &--    &--             & 28.53         & 14/22             \\ %\cline{2-9}
                                                & SurroundOcc\cite{wei2023surroundocc}  & ICCV 2023             &34.72& 7/14     &11.86 & 10/17         & --            & --                \\ %\cline{2-9}                      
                                                & OctreeOcc\cite{lu2023octreeocc}       & arXiv 2023            &44.71& 2/14     &13.12 &  5/17         & 44.02         &  8/22             \\ %\cline{2-9}
                                                & FlashOcc \cite{yu2023flashocc}        & arXiv 2023            &--   &--        &--    &--             & 45.51         &  7/22             \\ %\cline{2-9}
                                                & DepthSSC\cite{yao2023depthssc}        & arXiv 2023            &44.58& 3/14     & 13.11&  6/17         & --            & --                \\ %\cline{2-9}
                                                & Multi-Scale Occ\cite{ding2023multi}   & arXiv 2023            &--   &--        &--    &--             & 49.36         &  5/22             \\ %\cline{2-9}
                                                & FastOcc\cite{hou2024fastocc}          & ICRA 2024            &--   &--        &--    &--             & 39.21         & 11/22  
                                                \\ %\cline{2-9}
                                                &MonoOcc\cite{zheng2024monoocc}        & ICRA 2024            &--   &--        & 15.63&  2/17         & --            & --
                                                \\ %\cline{2-9}
                                                & PanoOcc\cite{wang2023panoocc}         & CVPR 2024            &--   &--        &--    &--             & 42.13         &  9/22 
                                                \\ %\cline{2-9}
                                                & SparseOcc\cite{liu2023fully}          & ECCV 2024            &--   &--        &--    &--             & 30.95         & 13/22     
                                                \\ \hline
                                                
\multirow{7}{*}{Label-efficient methods} 
& OVO\cite{tan2023ovo}                  & arXiv 2023            &--   &--        & 9.56 & 13/17         & --            & --                \\ %\cline{2-9}
& SelfOcc\cite{huang2023selfocc}        & arXiv 2023            &--   &--        &--    &--             & 9.30          & 21/22             \\ %\cline{2-9}
                                                & OccNeRF\cite{zhang2023occnerf}        & arXiv 2023            &--   &--        &--    &--             & 10.81         & 20/22             \\ %\cline{2-9}
                                                & RenderOcc\cite{pan2023renderocc}      & arXiv 2023            &--   &--        &12.87 &  7/17         & 26.11         & 17/22             \\ %\cline{2-9}
                                                & UniOcc\cite{pan2023uniocc}            & arXiv 2023            &--   &--        &--    &--             & 51.27         &  3/22             \\ %\cline{2-9}
                                                & RadOcc\cite{zhang2024radocc}          & AAAI 2024             &--   &--        &--    &--             & 49.98         &  4/22             \\ %\cline{2-9}
                                                & OccFlowNet\cite{boeder2024occflownet} & arXiv 2024            &--   &--        &--    &--             & 33.86         &  12/22             \\ \hline
\end{tabular}
%}
\label{table:table2}
\end{table*}

Since the current 3D occupancy annotations are generated from LiDAR points ground truth, there is only one layer of occupancy ground truth on the reflective surface for larger objects such as cars, while the interior is empty. This will have a great adverse impact on $IoU$ and $mIoU$ when the predicted depth has a small deviation. In the future, we should consider generating solid ground truth or matching occupied voxels within a certain range to improve this evaluation metric. Additionally, 3D occupancy prediction needs to predict whether each voxel is occupied and give the semantic label of the occupied voxel, requiring a lot of computational resources and memory. Future evaluations should also analyze parameters and inference time to compare the efficiency of different methods.

\subsection{Key challenges}

Although significant progress has been made in recent years for vision-based 3D occupancy prediction, it still faces limitations from feature representation, practical applications and annotation costs. For this task, there are three key challenges: (1) Obtaining perfect 3D features from 2D visual inputs is difficult. The goal of vision-based 3D occupancy prediction is to achieve detailed perception and comprehension of the 3D scene solely from image inputs, yet the inherent absence of depth and geometric information in images poses a significant challenge in directly learning 3D feature representations from them. (2) Heavy computational load in 3D space. 3D occupancy prediction typically requires representing the environmental space using 3D voxel features, which inevitably involves operations like 3D convolutions for feature extraction, substantially increasing computational and memory overhead and hindering practical deployment. (3) Expensive fine-grained annotation. 3D occupancy prediction involves predicting the occupancy status and semantic categories of high-resolution voxels, but achieving this typically demands fine-grained semantic annotation for each voxel, which is both time-consuming and costly, posing a bottleneck for this task.

In response to these key challenges, research work of vision-based 3D occupancy prediction for autonomous driving has gradually formed three main lines, i.e. feature enhancement, deployment-friendly and label-efficient. The feature enhancement method alleviates the discrepancy between 3D spatial output and 2D spatial input by optimizing the network's feature representation capability. Deployment-friendly methods aim to significantly reduce resource consumption while ensuring performance by designing concise and efficient network architectures. Label-efficient methods expect to achieve satisfactory performance even with insufficient or completely absent annotations. Next, we will provide a comprehensive overview of current methods around these three branches.

\section{Feature enhancement methods}
\label{FE}

The task of vision-based 3D occupancy prediction involves predicting the occupancy status and semantic information of 3D voxel space from 2D image space, posing a key challenge in obtaining perfect 3D features from 2D visual inputs. To address this issue, some methods improve occupancy prediction from the perspective of feature enhancement, including learning from Bird's Eye View (BEV), Tri-Perspective View (TPV), and 3D voxel representations.

\subsection{BEV-based methods}
\begin{figure}[!ht]
    \centering
    \includegraphics[width=\linewidth]{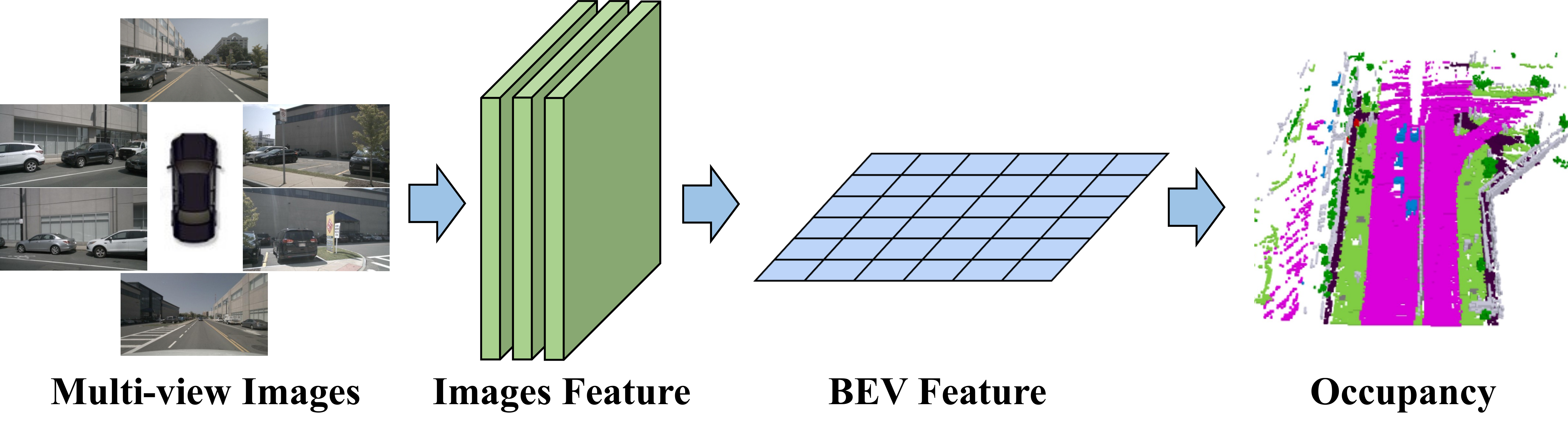}
    \caption{Illustration of BEV-based methods.}
    \label{fig:bev}
\end{figure}

An effective approach for learning occupancy is based on the Bird's Eye View (BEV), which offers features that are insensitive to occlusion and contain certain depth geometric information. By learning a strong BEV representation, robust 3D occupancy scene reconstruction can be achieved. Image features are first extracted from visual inputs using a 2D backbone network, followed by obtaining BEV features through viewpoint transformation, and ultimately completing 3D occupancy prediction based on the BEV feature representation. The illustration of BEV-based methods is shown in Figure \ref{fig:bev}.

One direct approach is to leverage BEV learning from other tasks, such as utilizing methods like BEVDet~\cite{huang2021bevdet} and BEVFormer~\cite{li2022bevformer} in 3D object detection. To extend these methods for occupancy learning, one can incorporate or replace an occupancy head during training to obtain the final results. This adaptation allows for the integration of occupancy estimation into existing BEV-based frameworks, enabling the simultaneous detection and reconstruction of 3D occupancy in a scene. Based on the powerful baseline BEVFormer, OccTransformer~\cite{liu2024occtransformer} employs data augmentation to increase the diversity of training data for improved model generalization and utilizes a strong image backbone to extract more informative features from input data. It also introduces a 3D Unet Head to better capture spatial information of scenes and incorporates additional loss functions for improved model optimization. 

\begin{figure}[!ht]
    \centering
    \includegraphics[width=\linewidth]{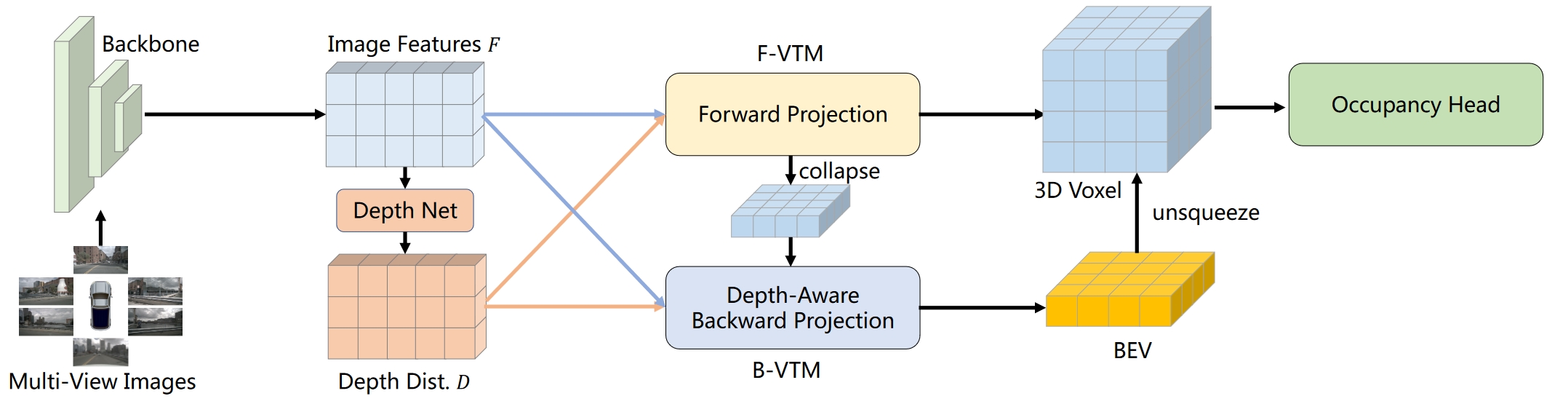}
    \caption{FB-OCC \cite{li2023fb2} applies forward and backword projection to generate dense BEV features, which are unsqueezed for 3D occupancy prediction.}
    \label{fig:fb-occ}
\end{figure}

View transformation is a crucial step in generating BEV representations, which involves transitioning between image features and BEV representations, with forward projection and backward projection being the most prominent paradigms. Forward projection lifts image features from 2D space to BEV space via LSS~\cite{philion2020lift} and weights the features based on depth, but tends to produce sparse BEV projections. Backward projection defines voxel positions in 3D space and projects them onto the 2D image plane to generate dense BEV features, but points along the projection rays acquire indistinctive identical features. FB-BEV~\cite{li2023fb} propose forward-backward projection, employing backward projection to refine necessary BEV grids to reduce sparsity and further introducing depth consistency into backward projection to assign different weights for each projection. Building upon the forward-backward projection BEV representation of FB-BEV, FB-OCC~\cite{li2023fb2} further explores novel designs and optimizations for the 3D occupancy prediction task, including joint depth-semantic pretraining, unified voxel-BEV representation, model amplification, and effective post-processing strategies such as test-time augmentation model ensemble. The overall architecture of FB-OCC is shown in Figure \ref{fig:fb-occ}.

\subsection{TPV-based methods}
\begin{figure}[!ht]
    \centering
    \includegraphics[width=\linewidth]{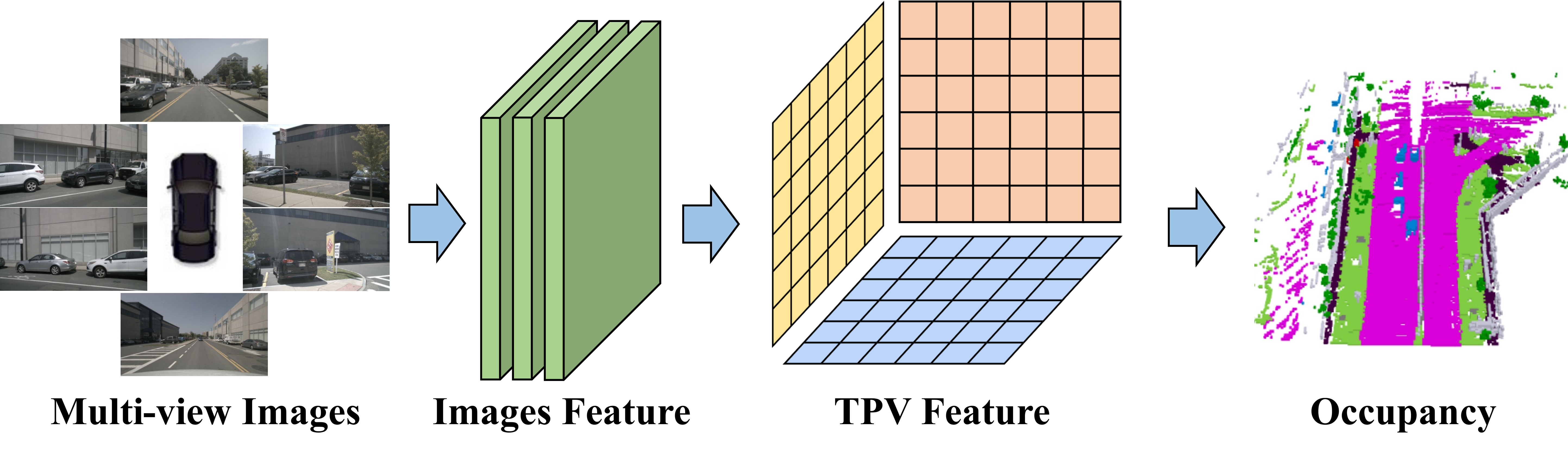}
    \caption{Illustration of TPV-based methods.}
    \label{fig:tpv0}
\end{figure}

%zjq
While BEV-based representations have certain advantages compared to images, as they essentially provide a top-down projection of 3D space, they inherently lack the ability to describe the fine-grained 3D structure of scenes using only a single plane. Tri-Perspective View (TPV) based methods utilize three orthogonal projection planes to model the 3D environment, further enhancing the representation capability of visual features for occupancy prediction. First, image features are extracted from visual inputs using a 2D backbone network. Subsequently, these image features are elevated to a three-view space, and ultimately, 3D occupancy prediction is accomplished based on the feature representation from the three projection viewpoints. The illustration of TPV-based methods is shown in Figure \ref{fig:tpv0}.

\begin{figure}[!ht]
    \centering
    \includegraphics[width=\linewidth]{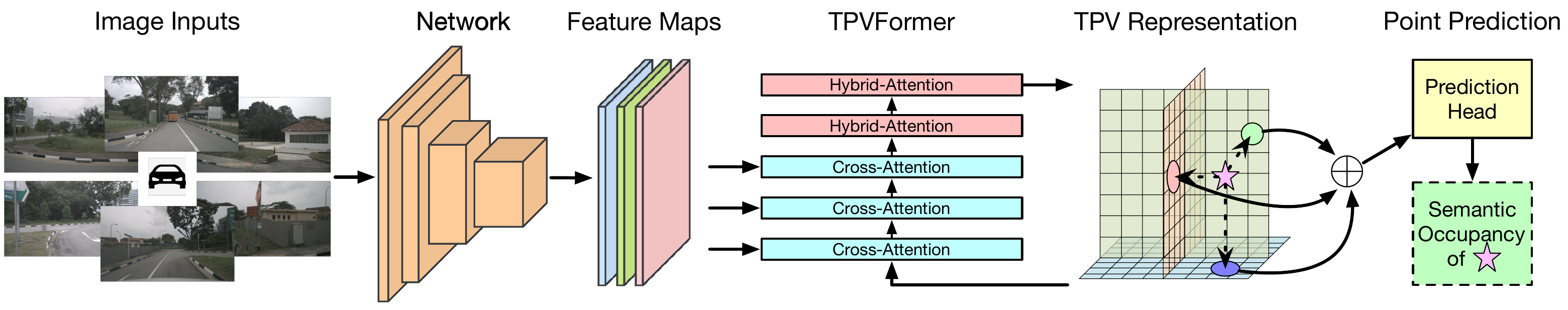}
    \caption{TPVFormer \cite{huang2023tri} generates Tri-Perspective View features, which are sampled to get the 3D features for occupancy prediction.}
    \label{fig:tpv1}
\end{figure}

Apart from BEV features, TPVFormer~\cite{huang2023tri} also generates the features in the front view and side view in the same way. The triple views are orthogonal to each other and squeeze the 3D space information along different axes. The whole 3D structure of the driving scene can be recovered by combining the TPV features. To retrieve the feature of a certain position in the 3D space from TPV features, the 3D coordinates are first projected on each of the views. After that, bilinear sampling is utilized to obtain the features from the triple views, which are summarized as the composite features. This mechanism makes 2D TPV features able to flexibly recover the 3D space without losing too much information. In practice, the TPV features are generated by utilizing a Transformer-based encoder, which consists of several cross-attention between TPV grid queries and 2D image features. Finally, the cross-view hybrid attention among TPV features achieves interaction between the three planes, improving the capability of the TPV feature to represent the whole scene. The overall architecture of TPVFormer is shown in Figure \ref{fig:tpv1}.

Considering TPVFormer only focuses on spatial cues while neglecting temporal cues, S2TPVFormer~\cite{silva2024s2tpvformer} further extends the temporal fusion ability of TPVFormer. It adopts TPV as a latent ego spatial representation and proposes Temporal Cross-View Hybrid Attention, which first interacts with the current BEV features with previous BEV features. After that, the front and side view features will interact with the updated BEV features to fusion temporal information. Taking into account the non-uniform distribution of LiDAR point clouds in space, with denser points nearby and sparser points further away, PointOcc~\cite{zuo2023pointocc} proposes establishing a three-view representation of point clouds in cylindrical coordinates to achieve finer-grained modeling of nearby regions, and employing spatial grouping pooling during the projection process to better preserve 3D structural information. Specifically, given the input point cloud, it is first transformed into tri-view features through voxelization and spatial pooling operations in cylindrical coordinates. Then, a 2D image backbone is employed to encode each tri-view feature plane, and multi-scale features from each plane are aggregated to decode high-resolution tri-view feature representations. For any query point in 3D space, it is projected onto each tri-view plane through cylindrical coordinate transformation to sample the corresponding features. These features from the tri-view planes are then aggregated using summation to form the final 3D feature representation.

\subsection{Voxel-based methods} 
\begin{figure}[!ht]
    \centering
    \includegraphics[width=\linewidth]{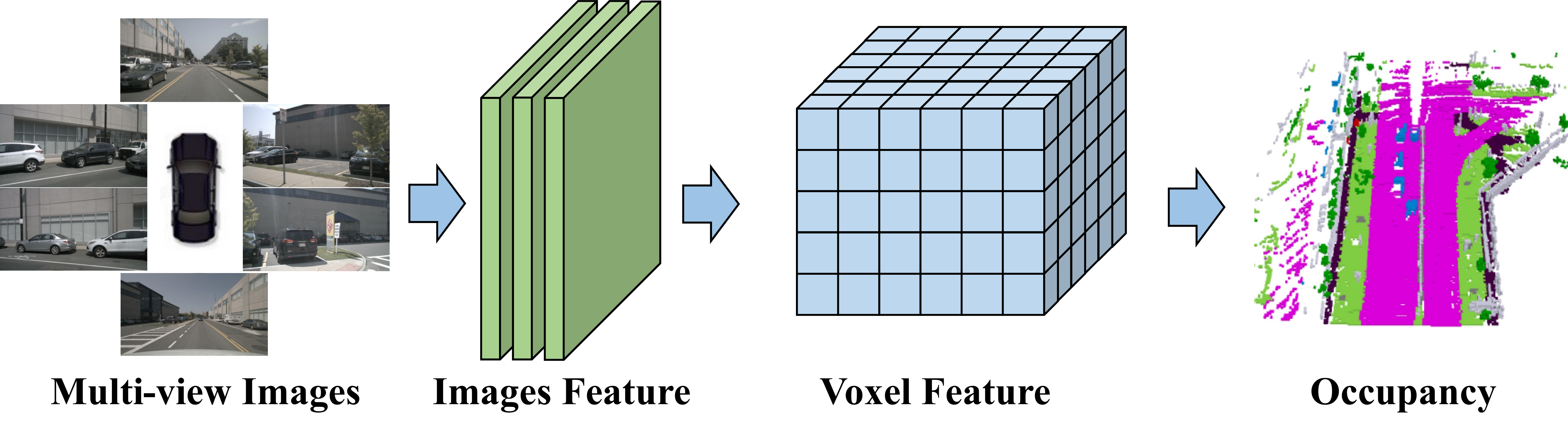}
    \caption{Illustration of Voxel-based methods.}
    \label{fig:voxel}
\end{figure}

Apart from transforming 3D space into projection perspective such as BEV or TPV, there are also methods that directly operate on 3D voxel representations. One key advantage of these approaches is their ability to learn directly from the raw 3D space, minimizing information loss. By leveraging the original 3D voxel data, these methods can effectively capture and utilize complete spatial information, leading to a more precise and comprehensive understanding of occupancy. Firstly, image features are extracted using a 2D backbone network. Then, a specially designed mechanism based on convolution is utilized to bridge the 2D and 3D representations, or a query-based approach is employed to directly obtain the 3D representation. Finally, based on the learned 3D representation, a 3D occupancy head is utilized to complete the final prediction. The illustration of Voxel-based methods is shown in Figure \ref{fig:voxel}.

\paragraph{Convolution-based methods} 
One way is to utilize specially designed convolutional architectures to bridge the gap from 2D to 3D and learn 3D occupancy representations. A prominent example of this approach is adopting the U-Net~\cite{ronneberger2015u} architecture as the carrier for feature bridging. The U-Net architecture employs an encoder-decoder structure with skip connections between the upsampling and downsampling paths, preserving both low-level and high-level feature information to alleviate information loss. With convolutional layers at different depths, the U-Net structure can extract features at various scales, aiding the model in capturing both local details and global contextual information in images, thereby enhancing the model's understanding of complex scenes for effective occupancy prediction.

\begin{figure}[!ht]
    \centering
    \includegraphics[width=\linewidth]{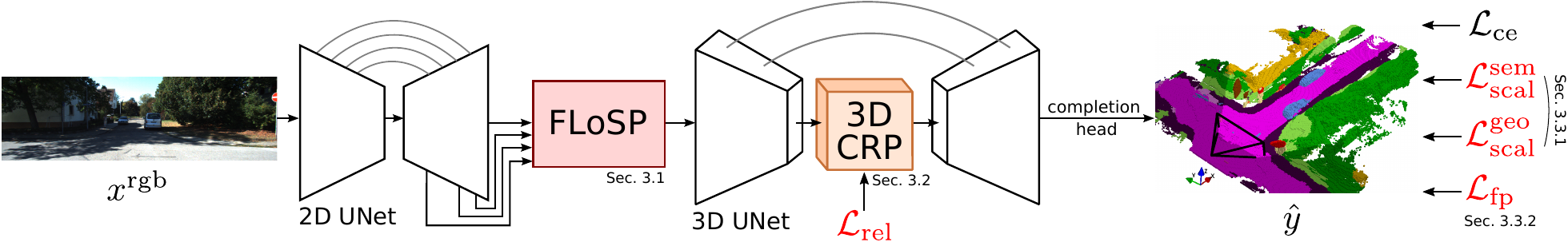}
    \caption{Monoscene~\cite{cao2022monoscene} leverages 2D and 3D UNets, bridged by the Features Line of Sight Projection (FLoSP), and a 3D Context Relation Prior (3D CRP) to enhance spatio-semantic awareness.}
    \label{fig:Monoscene}
\end{figure}

Monoscene~\cite{cao2022monoscene} utilizes U-net for vision-based 3D occupancy prediction. It introduces a mechanism called 2D Features Line of Sight Projection (FLoSP), which utilizes feature perspective projection to project 2D features onto the 3D space, and calculates the coordinates of each point in the 3D feature space on the 2D feature based on the imaging principle and camera parameters for sampling the features in the 3D feature space.
This approach lifts 2D features to a unified 3D feature map
and serves as a crucial component in connecting 2D and 3D U-nets.  Monoscene also proposes a 3D Context Relation Prior (3D CRP) layer, inserted at the 3D UNet bottleneck, which learns n-way voxel-to-voxel semantic scene-wise relation maps. This provides the network with a global receptive field, and increases spatio-semantic awareness due to the relations discovery mechanism. The overall architecture of Monoscene is shown in Figure \ref{fig:Monoscene}. 

Furthermore, InverseMatrixVT3D~\cite{ming2024inversematrixvt3d} achieves occupancy prediction through matrix multiplications and convolutions. Specifically, this method initially feeds the surround view into a 2D backbone network to extract image features across multiple scales. Subsequently, a multi-scale global-local projection module is employed to perform matrix multiplication between the feature maps and projection matrices for constructing multi-scale 3D volume features and BEV planes. Each level of 3D volume feature and BEV plane applies a global-local attention fusion module to obtain the final 3D volume representation. Finally, 3D deconvolution is applied for upsampling to enable skip connections at each level, with supervised signals also applied at each level. MiLO~\cite{myeongjin2023milo} uses the 3D version of the ResNet~\cite{he2016deep} and FPN~\cite{lin2017feature} to process the voxel features and suppress the ambiguity within the feature localization. The core idea of SGN\cite{mei2023camera} is to leverage geometric priors and occupancy information to propagate semantics from semantic and occupancy-aware seed voxels to the entire scene, utilizing anisotropic convolutions to achieve flexible receptive fields. In contrast to traditional methods, SGN adopts a dense-sparse-dense design and introduces mixed guidance and effective voxel aggregation to enhance intra-class feature separation and accelerate the convergence of semantic diffusion.

\paragraph{Query-based methods}
Another way to learning from 3D space involves generating a set of queries to capture the scene's representation. In this method, query-based techniques are used to produce query proposals, which are then employed to learn a comprehensive representation of the 3D scene. Subsequently, cross-attention on images and self-attention mechanisms are applied to refine and enhance the learned representation. This approach not only enhances the understanding of the scene but also enables accurate reconstruction and occupancy prediction in the 3D space. Additionally, query-based methods offer greater flexibility to adjust and optimize based on different data sources and query strategies, enabling better capture of both local and global contextual information that facilitates 3D occupancy prediction representations.

\begin{figure}[!ht]
    \centering
    \includegraphics[width=\linewidth]{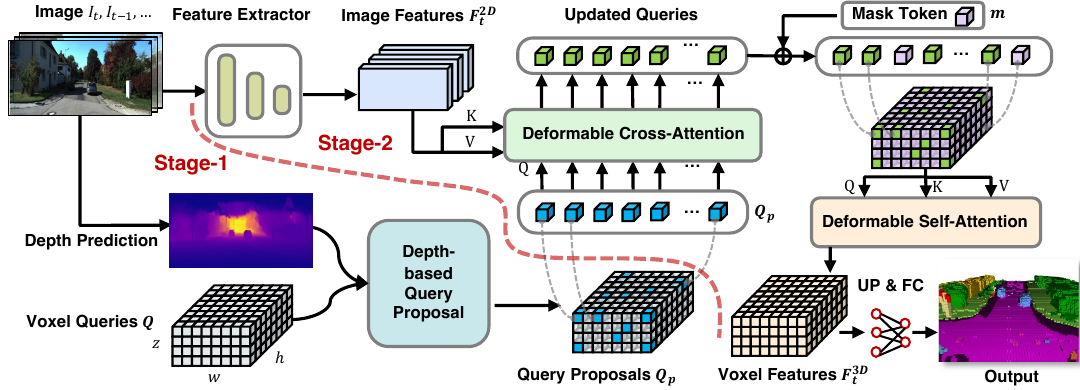}
    \caption{VoxFormer \cite{li2023voxformer} uses estimated depth as a prior to help retrieve the relevant features.}
    \label{fig:voxformer}
\end{figure}

Depth can serve as a valuable prior for selecting occupied queries~\cite{miao2023occdepth, yao2023depthssc, li2023voxformer}, in Voxformer~\cite{li2023voxformer}, estimated depth is leveraged as a prior to predict occupancy and select relevant queries. Only the occupied queries are used to gather information from images using deformable attention~\cite{zhu2020deformable}. The updated query proposals and masked tokens are then combined to reconstruct the voxel features. Voxformer extracts 2D features from RGB images and then utilizes a set of sparse 3D voxel queries to index these 2D features, linking 3D positions to the image stream using the camera projection matrix. Specifically, voxel queries are learnable parameters of the 3D grid shape, intended to query features from the image into the 3D volume using attention mechanisms. The entire framework is a two-stage cascade consisting of class-agnostic proposals and class-specific segmentation. Stage-1 generates class-agnostic query proposals, while Stage-2 employs an architecture similar to MAE~\cite{he2022masked} to propagate information to all voxels. Finally, voxel features are upsampled for semantic segmentation. The overall architecture of VoxFormer is shown in Figure~\ref{fig:voxformer}. 

\begin{figure}[!ht]
    \centering
    \includegraphics[width=\linewidth]{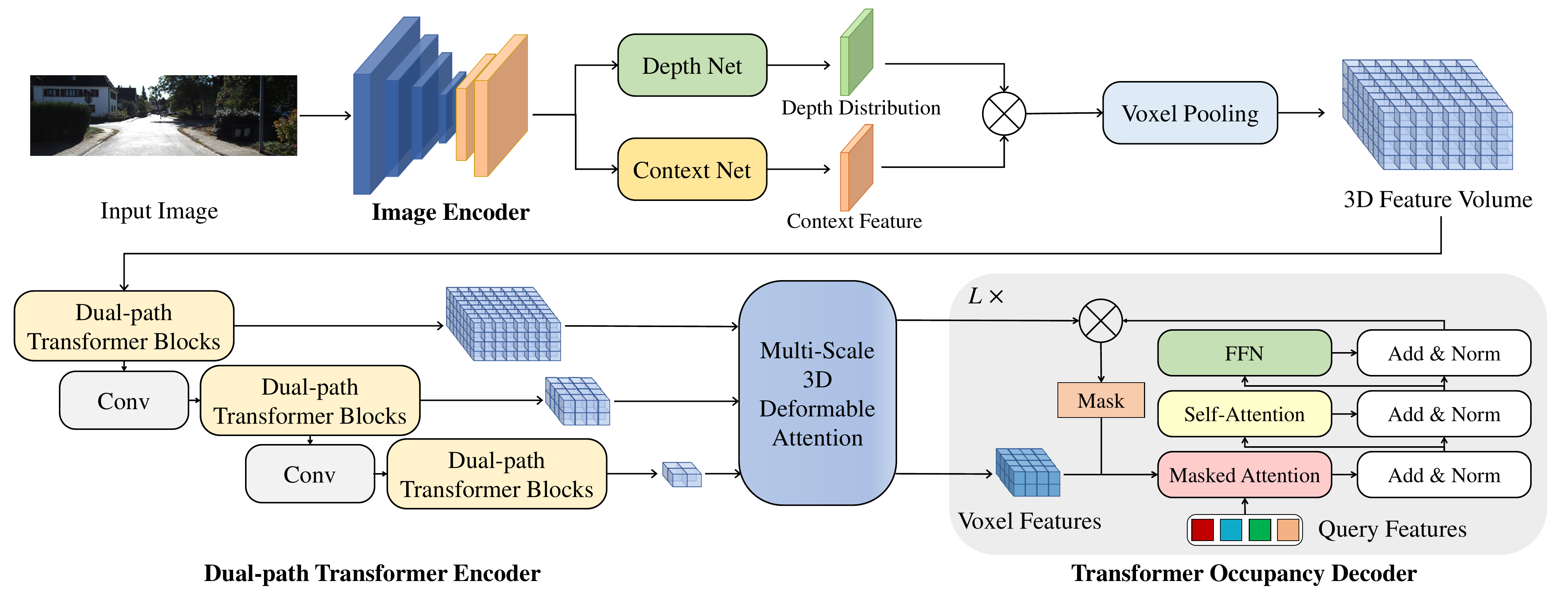}
    \caption{OccFormer \cite{zhang2023occformer} encodes 3D voxels in different directions by utilizing dual-path Transformer.}
    \label{fig:occformer}
\end{figure}

OccFormer~\cite{zhang2023occformer}, on the other hand, proposes a dual-path Transformer to encode 3D voxels in different directions. It is also the first method to utilize the mask classification model Mask2Former~\cite{cheng2022masked} for 3D semantic occupancy prediction. The Dual-path Transformer encoder effectively captures fine-grained details and scene-level layout through both local and global pathways. The local path operates along each 2D BEV slice, using shared-window attention to capture fine-grained details, while the global path processes the collapsed BEV features to gain scene-level understanding. Finally, the outputs from both pathways are adaptively fused to generate the output's 3D feature volume. For the occupancy decoder, considering the inherent sparsity and class imbalance, the proposed preserve-pooling computes attention on masked regions for better preserve rare classes and the class-guided sampling is employed to capture foreground regions, significantly improving performance. The overall architecture of OccFormer is shown in Figure~\ref{fig:occformer}. 

Additionally, Symphonies~\cite{jiang2023symphonize} presents a novel paradigm that integrates instance queries to orchestrate 2D-to-3D reconstruction and 3D scene modeling, dynamically encoding instance-centric semantics for intricate interactions between image-based and volumetric domains and enabling holistic scene comprehension through the efficient fusion of instance queries to alleviate geometric ambiguities such as occlusion and perspective errors. COTR~\cite{ma2023cotr} comprises a geometry-aware occupancy encoder and a semantic-aware group decoder for reconstructing a compact 3D occupancy representation. The geometry-aware occupancy encoder generates a compact geometric occupancy feature through explicit-implicit view transformation, mitigating the sparsity of occupancy features while retaining sufficient geometric information. The semantic-aware group decoder partitions ground truth labels into multiple groups according to semantic granularity and sample count, creating corresponding mask queries for each semantic group, and trains the network through group-wise one-to-many assignment. This grouping strategy balances supervision signals and mitigates suppression from common to rare objects. UniVision~\cite{hong2024univision} unifies two key tasks in vision-centric 3D perception: occupancy prediction and object detection. Specifically, it proposes an explicit-implicit view transformation module for complementary 2D-3D feature transformation, a local-global feature extraction and fusion module for efficient and adaptive voxel and BEV feature interaction, a joint data augmentation strategy and a progressive loss weight adjustment strategy for enhancing the efficiency and stability of multi-task framework training.

\begin{table*}[!ht]
    \caption{Performance of the Feature enhancement methods on Occ3D-nuScenes dataset. The best one is highlighted in \textbf{bold}. ``Rank'' denotes the accuracy rank among all methods. Note that some experimental settings may not be exactly the same, and the results are just collected here for reference.}
    \renewcommand{\tabcolsep}{3pt}
    \centering
    \scalebox{0.85}{
    	\begin{tabular}{l|l|ccccccccccccccccc|c|c}
		\toprule
		Category & Method
        & \rotatebox{90}{\textcolor{nothers}{$\blacksquare$} others}
		& \rotatebox{90}{\textcolor{nbarrier}{$\blacksquare$} barrier}
		& \rotatebox{90}{\textcolor{nbicycle}{$\blacksquare$} bicycle}
		& \rotatebox{90}{\textcolor{nbus}{$\blacksquare$} bus}
		& \rotatebox{90}{\textcolor{ncar}{$\blacksquare$} car}
		& \rotatebox{90}{\textcolor{nconstruct}{$\blacksquare$} const. veh.}
		& \rotatebox{90}{\textcolor{nmotor}{$\blacksquare$} motorcycle}
		& \rotatebox{90}{\textcolor{npedestrian}{$\blacksquare$} pedestrian}
		& \rotatebox{90}{\textcolor{ntraffic}{$\blacksquare$} traffic cone}
		& \rotatebox{90}{\textcolor{ntrailer}{$\blacksquare$} trailer}
		& \rotatebox{90}{\textcolor{ntruck}{$\blacksquare$} truck}
		& \rotatebox{90}{\textcolor{ndriveable}{$\blacksquare$} drive. suf.}
		& \rotatebox{90}{\textcolor{nother}{$\blacksquare$} other flat}
		& \rotatebox{90}{\textcolor{nsidewalk}{$\blacksquare$} sidewalk}
		& \rotatebox{90}{\textcolor{nterrain}{$\blacksquare$} terrain}
		& \rotatebox{90}{\textcolor{nmanmade}{$\blacksquare$} manmade}
		& \rotatebox{90}{\textcolor{nvegetation}{$\blacksquare$} vegetation}
        & mIoU & Rank
		\\
		\midrule
        \multirow{6}{*}{BEV-based} 
        & BEVDet\cite{huang2021bevdet} & 4.39 &30.31 &0.23& 32.26& 34.47 &12.97& 10.34 &10.36& 6.26& 8.93& 23.65& 52.27& 24.61& 26.06& 22.31& 15.04& 15.10 & 19.38 & 19 \\
        & BEVFormer\cite{li2022bevformer} & 5.85 & 37.83 & 17.87 & 40.44 & 42.43 & 7.36 & 23.88 & 21.81 & 20.98 & 22.38 & 30.70 & 55.35 & 28.36 & 36.00 & 28.06 & 20.04 & 17.69 & 26.88 & 16 \\
        & BEVStereo\cite{li2023bevstereo} & 12.15 & 49.63 & 25.10 & 52.02 & 54.46 & 27.87 & 27.99 & 28.94 & 27.23 & 36.43 & 42.22 & 82.31 & 43.29 & 54.46 & 57.90 & 48.61 & 43.55 & 42.02& 10 \\
        & OccTransformer\cite{liu2024occtransformer} & 26.91 & 53.57 & 39.53 & 47.56 & 59.54 & 32.59 & 44.34 & 37.36 & 37.28 & 54.81 & 44.7 & 84.61 & 55.15 & 60.34 & 56.35 & 57.14 & 45.04 & 49.23 & 6 \\
        & FB-OCC\cite{li2023fb} & \textbf{28.28} & \textbf{56.70} & \textbf{44.35} & 51.37 & \textbf{61.81} & 35.12 & 47.38 & 41.56 & 39.88 & \textbf{57.96} & \textbf{48.39} & 86.66 & 56.97 & \textbf{64.66} & 61.23 & 62.78 & 52.35 & \textbf{52.79} & 1 \\
        
        \midrule
        \multirow{1}{*}{TPV-based} & TPVFormer\cite{huang2023tri} & 6.67 & 39.20 & 14.24 & 41.54 & 46.98 & 19.21 & 22.64 & 17.87 & 14.54 & 30.20 & 35.51 & 56.18 & 33.65 & 35.69 & 31.61 & 19.97 & 16.12 & 28.34 & 15 \\

        \midrule
        \multirow{6}{*}{Voxel-based} & MonosSene\cite{cao2022monoscene} & 1.75 & 7.23 & 4.26 & 4.93 & 9.38 & 5.67 & 3.98 & 3.01 & 5.90 & 4.45 & 7.17 & 14.91 & 6.32 & 7.92 & 7.43 & 1.01 & 7.65 & 6.06 & 22 \\
        & CTFOcc\cite{tian2023occ3d} & 8.09 & 39.33 & 20.56 & 38.29 & 42.24 & 16.39 & 24.52 & 22.72 & 21.05 & 22.98 & 31.11 & 53.33 & 33.84 & 37.98 & 33.23 & 20.79 & 18.00 & 28.53 & 14\\
        & MiLO\cite{myeongjin2023milo} & 27.80 & 56.28 & 42.62 & 50.27 & 61.01 & \textbf{35.41} & \textbf{47.97} & \textbf{38.90} & \textbf{40.29} & 56.66 & 47.03 & \textbf{86.96} & \textbf{57.48} & 63.64 & \textbf{62.53} & \textbf{63.00} & \textbf{53.74} & 52.45 & 2 \\
        & OccFormer\cite{zhang2023occformer} & 5.94 & 30.29 & 12.32 & 34.40 & 39.17 & 14.44 & 16.45 & 17.22 & 9.27 & 13.90 & 26.36 & 50.99 & 30.96 & 34.66 & 22.73 & 6.76 & 6.97 & 21.93 & 18 \\
        & PanoOcc\cite{wang2023panoocc} & 11.67 & 50.48 & 29.64 & 49.44 & 55.52 & 23.29 & 33.26 & 30.55 & 30.99 & 34.43 & 42.57 & 83.31 & 44.23 & 54.40 & 56.04 & 45.94 & 40.40 & 42.13  & 9\\
        & OctreeOcc\cite{lu2023octreeocc} & 11.96 & 51.70 & 29.93 & \textbf{53.52} & 56.77 & 30.83 & 33.17 & 30.65 & 29.99 & 37.76 & 43.87 & 83.17 & 44.52 & 55.45 & 58.86 & 49.52 & 46.33 & 44.02 &  8 \\
        
		\bottomrule
	\end{tabular}
    }
    \label{table:feature_enhance_methods}
\end{table*}

The performance comparison of feature enhancement methods on the Occ3D-nuScenes dataset is shown in Table~\ref{table:feature_enhance_methods}. The results demonstrate that approaches that directly process the voxel representations are generally able to achieve strong performance, as they do not suffer from significant information loss during the computation.
Additionally, although BEV-based methods only have a single projected viewpoint for feature representation, they can still achieve comparable performance due to the rich information contained in the Bird's Eye View and their insensitivity to occlusion and scale variations. Furthermore, by reconstructing the 3D information from multiple complementary views, the method based on Tri-Perspective View (TPV) is able to mitigate potential geometric ambiguities and capture a more comprehensive understanding of the scene, leading to effective 3D occupancy prediction. It is worth noting that FB-OCC~\cite{li2023fb} simultaneously utilizes forward and backward view transformation modules, allowing them to enhance each other to obtain higher quality BEV representations, and has also achieved excellent performance. This indicates that through effective feature enhancement, BEV-based methods also have great potential to improve 3D occupancy prediction.

\section{Deployment-friendly methods}
\label{CF}

Due to its extensive scope and complex data nature, directly learning occupancy representations from 3D space is highly challenging. The high dimensionality and intensive computation associated with 3D voxel representations make the learning process expensive resource-demanding, which is unfavorable for practical deployment applications. Therefore, methods that design deployment-friendly 3D representations aim to reduce computational costs and improve learning efficiency. This section introduces approaches to address the computational challenges in 3D scene occupancy estimation, focusing on developing accurate and efficient methods rather than directly processing the entire 3D space. The discussed techniques include perspective decomposition and coarse-to-fine refinement, which have been demonstrated in recent works to enhance the computational efficiency of 3D occupancy prediction.

\subsection{Perspective decomposition methods}

By separating viewpoint information from 3D scene features or projecting them into a unified representation space, computational complexity can effectively be reduced, making the model more robust and generalizable. The core idea of this approach is to decouple the representation of three-dimensional scenes from viewpoint information, thereby reducing the number of variables to consider during feature learning and lowering computational complexity. Decoupling viewpoint information allows the model to generalize better, adapting to different viewpoint transformations without the need to relearn the entire model.

To address the computational burden of learning from the entire 3D space, a common approach is to employ Bird's-Eye View (BEV)~\cite{huang2021bevdet,li2022bevformer ,hou2024fastocc, yu2023flashocc} and  Tri-Perspective View (TPV)~\cite{huang2023tri, zuo2023pointocc, silva2024s2tpvformer} representations. By decomposing the 3D space into these separate view representations, the computational complexity is significantly reduced while still capturing essential information for occupancy prediction. The key idea is to first learn from the BEV and TPV perspectives, and then recover the full 3D occupancy information by combining the insights gained from these different views. This perspective decomposition strategy allows for more efficient and effective occupancy estimation compared to directly learning from the entire 3D space.

\begin{figure}[!ht]
    \centering
    \includegraphics[width=\linewidth]{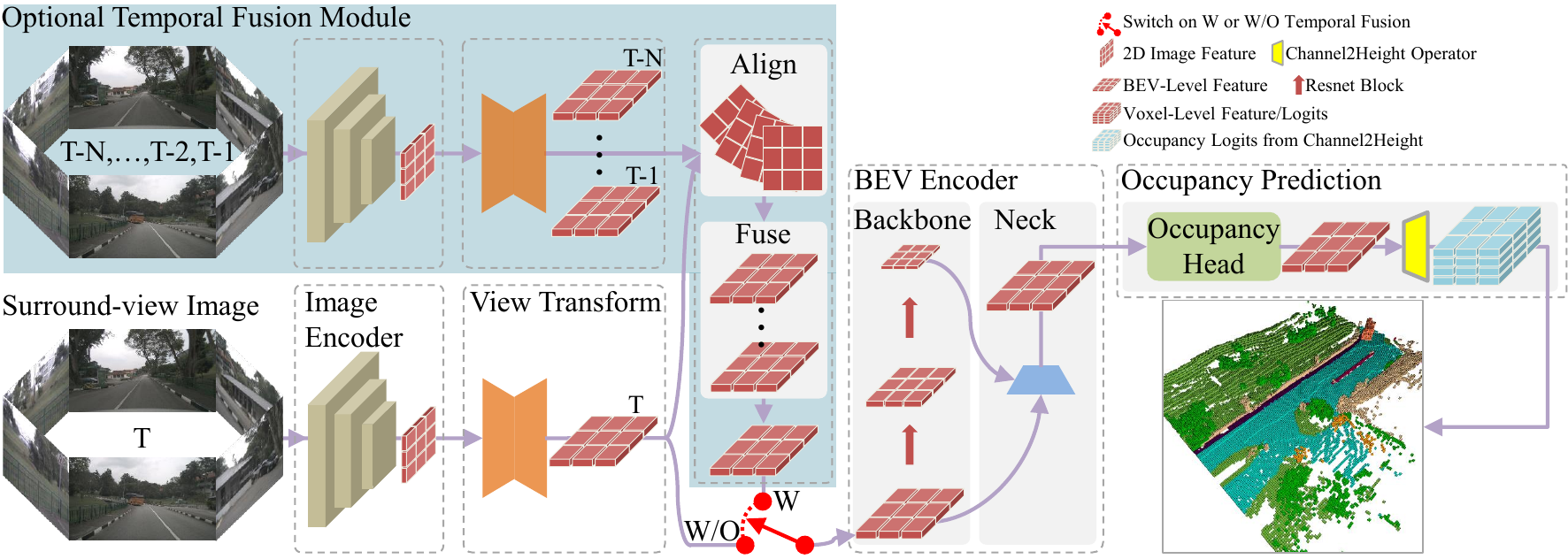}
    \caption{FlashOcc~\cite{yu2023flashocc} replaces 3D-Conv with 2D-Conv and employs channel-to-height transformation to replace the occupancy logits derived from 3D-Conv, achieving fast and memory-efficient occupancy prediction.}
    \label{fig:flashOcc}
\end{figure} 

FlashOcc~\cite{yu2023flashocc} aims to achieve fast and memory-efficient occupancy prediction. It directly replaces 3D convolutions in voxel-based occupancy methods with 2D convolutions, and combines channel-to-height transformation to reshape flattened BEV features into occupancy logits. FlashOcc is the first to apply the Channel-to-Height paradigm to occupancy tasks, specifically leveraging BEV-level features, completely avoiding the use of 3D deformable convolutions or transformer modules, and successfully preserving height information. As shown in Figure~\ref{fig:flashOcc}, FlashOcc comprises five basic modules: (1) A 2D image encoder responsible for extracting image features from multi-camera images. (2) A view transformation module responsible for mapping 2D perceptual view image features to a BEV representation. (3) A BEV encoder responsible for processing BEV feature information. (4) An occupancy prediction module responsible for predicting segmentation labels for each voxel. (5) An optional temporal fusion module aimed at integrating historical information to improve performance. FastOcc~\cite{hou2024fastocc} replaces the time-consuming 3D convolutional network with a novel residual-style architecture, predominantly processing features through a lightweight 2D BEV convolutional network, and supplementing them with integrated 3D Voxel features interpolated from original image features, thus efficiently obtaining complete 3D Voxel information.

\subsection{Coarse-to-fine methods}
Learning high-resolution fine-grained global voxel features directly from a large-scale 3D space is time-consuming and challenging. Therefore, some methods have begun to explore adopting a coarse-to-fine feature learning paradigm. Specifically, the network initially learns a coarse representation from images and subsequently refines and recovers the fine-grained representation of the whole scene. This two-step process aids in achieving a more accurate and effective prediction of the scene's occupancy.

\begin{figure}[!ht]
    \centering
    \includegraphics[width=\linewidth]{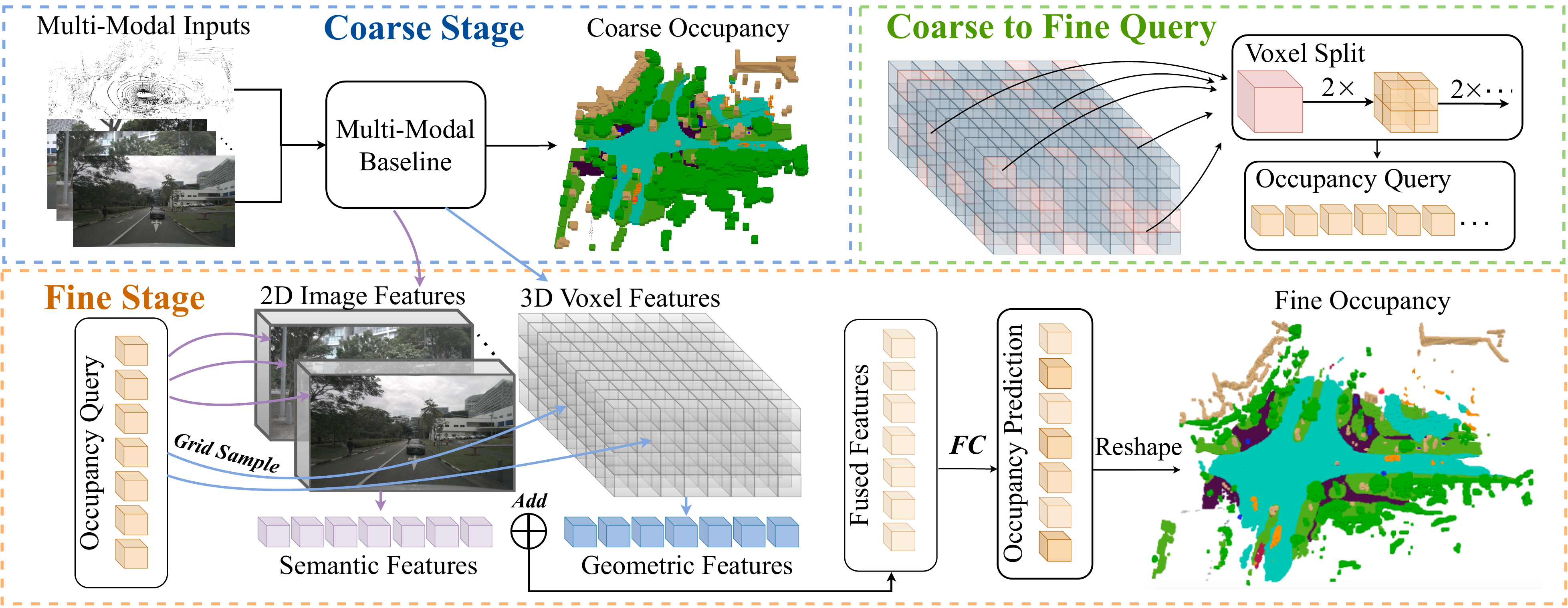}
    \caption{OpenOccupancy~\cite{wang2023openoccupancy} uses low-resolution queries to save on computation and then focuses only on the occupied voxels during refinement.}
    \label{fig:OpenOccupancy}
\end{figure} 

OpenOccupancy~\cite{wang2023openoccupancy} takes a two-step approach to learn occupancy representation in 3D space. As shown in Figure~\ref{fig:OpenOccupancy}, it first constructs queries with low resolutions to predict coarse occupancy. Then, it refines the results by focusing only on occupied voxels, improving the accuracy of the occupancy estimation. A Cascade Occupancy Network (CON) is adopted to efficiently handle the large number of empty voxels while providing accurate predictions.  SurroundOcc~\cite{wei2023surroundocc} first constructs image features at different levels and utilizes 2D-3D attention to elevate low-resolution high-level semantic features to a coarse-grained 3D voxel space. Then, deconvolution is applied to refine the granularity of voxels, avoiding the direct interaction of time-consuming 2D and 3D information at the highest-resolution layer.

\begin{figure}[!ht]
    \centering
    \includegraphics[width=\linewidth]{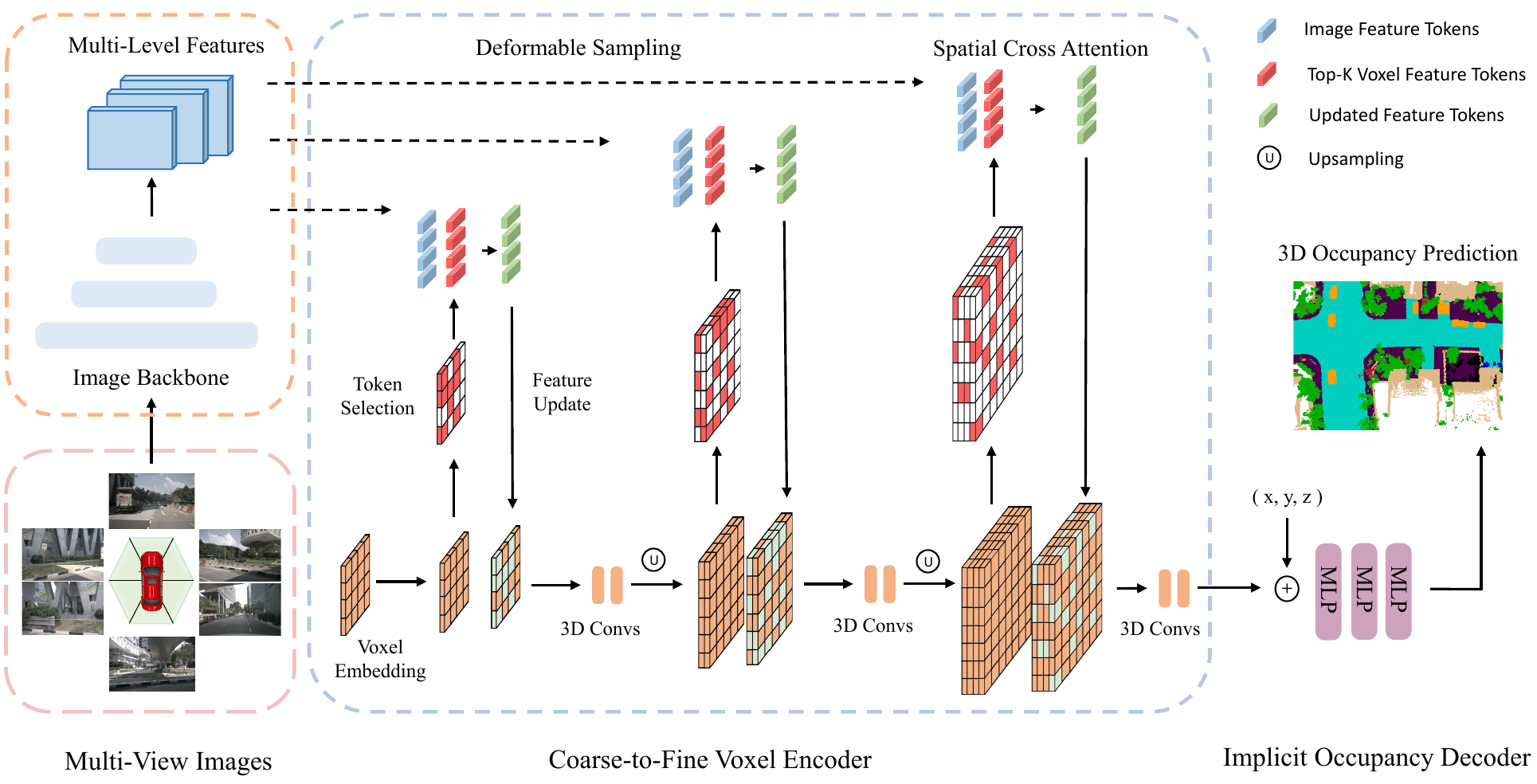}
    \caption{Occ3D~\cite{tian2023occ3d} utilizes a pyramid voxel encoder to gradually enhance voxel feature representation through coarse-to-fine incremental token selection and spatial cross-attention, strengthening spatial resolution and refining geometric details of objects.}
    \label{fig:Occ3D}
\end{figure}

Predicting 3D occupancy demands detailed geometric representations, and utilizing all 3D voxel tokens for interacting with ROIs in multi-view images would incur significant computational and memory costs. As shown in Figure~\ref{fig:Occ3D}, Occ3D~\cite{tian2023occ3d} proposes an incremental token selection strategy, selectively choosing foreground and uncertain voxel tokens during cross-attention computation, thus achieving adaptive and efficient computation without sacrificing accuracy. Specifically, at the beginning of each pyramid layer, each voxel token is inputted into a binary classifier to predict whether the voxel is empty, supervised by a binary ground truth occupancy map to train the classifier. 
PanoOcc~\cite{wang2023panoocc} proposes a seamless integration of object detection and semantic segmentation within a joint learning framework, fostering a more comprehensive understanding of the 3D environment. The method utilizes voxel queries to aggregate spatiotemporal information from multiple frames and multi-view images, merging feature learning and scene representation into a unified occupancy representation. Furthermore, it explores the sparsity nature of 3D space by introducing an occupancy sparsity module that gradually sparsifies occupancy during the coarse-to-fine upsampling process, significantly enhancing memory efficiency.

\begin{figure}[!ht]
    \centering
    \includegraphics[width=\linewidth]{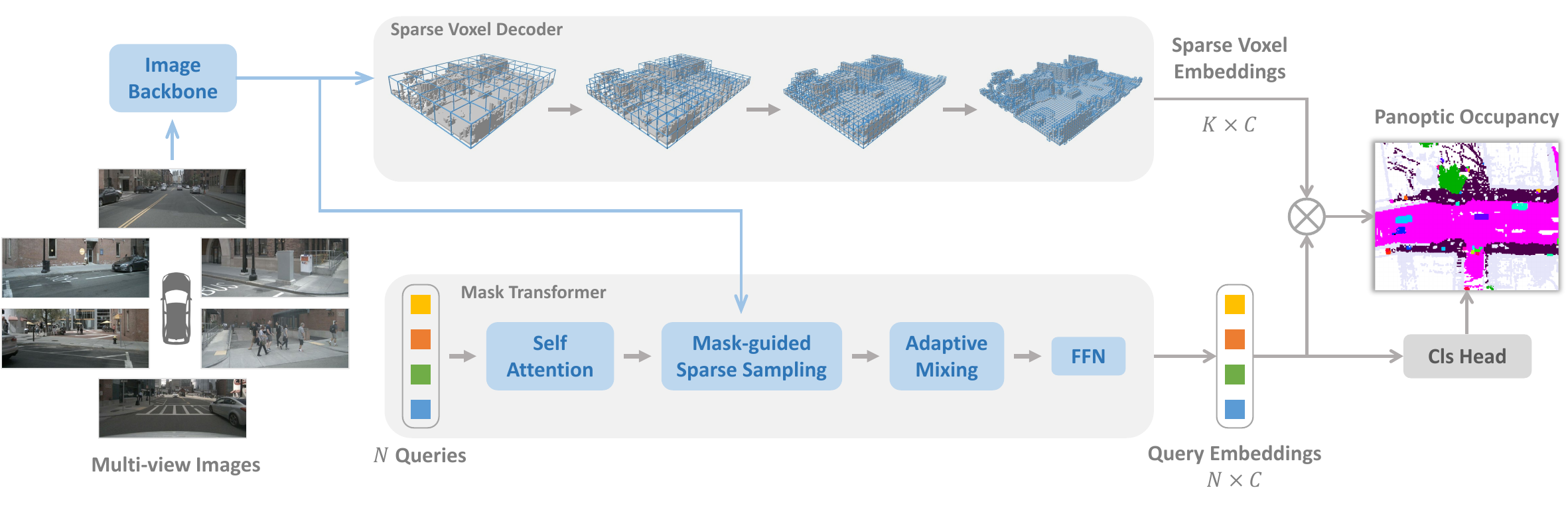}
    \caption{SparseOcc~\cite{liu2023fully} adopts fully sparse architecture to achieve high efficiency in 3D occupancy prediction.}
    \label{fig:sparseocc}
\end{figure}

Furthermore, certain approaches prioritize the utilization of fewer but more impactful queries to decrease computational demands. 
SparseOcc~\cite{liu2023fully} introduces a sparse voxel decoder to reconstruct the sparse geometric structure of the scene, modeling only the non-empty regions, thereby significantly saving computational resources. Furthermore, it proposes mask-guided sparse sampling, utilizing sparse instance queries to predict masks and labels for each object in the sparse space. As shown in Figure~\ref{fig:sparseocc}, SparseOcc combines both sparse features, forming a fully sparse architecture that does not rely on dense 3D features or sparse-to-dense global attention operations. OctreeOcc~\cite{lu2023octreeocc} utilizes octree structures to provide varying modeling granularity for different regions, adapting to capture valuable information in 3D space. This approach reduces the number of voxels needed for modeling while retaining spatial information. Additionally, it introduces a semantic-guided octree initialization module and an iterative refinement module, leveraging semantic information to initialize  and iteratively refine the octree structure. Multi-Scale Occ~\cite{ding2023multi} processes multi-scale image features to generate multi-scale voxel features and predicts coarse-to-fine 3D occupancy. On the hand, OccupancyM3D~\cite{peng2023learning} proposes a method to predict occupancy labels in frustum space.

\begin{table*}[!ht]
    \caption{Performance of the Deployment-friendly methods on Occ3D-nuScenes dataset. Note that some experimental settings may not be exactly the same, and the results are just collected here for reference.}
    \renewcommand{\tabcolsep}{7pt}
    \centering
    \scalebox{0.9}{
    	\begin{tabular}{l|l|c|c|c|c|c|c|c}
		\toprule
		Category & Method & Backbone & Image Size & Platform & Latency & Memory & mIoU & Rank\\
		\midrule
        \multirow{7}{*}{\makecell[l]{Perspective \\ Decomposition}} 
        & BEVFormer\cite{li2022bevformer} & ResNet-101 & 640$\times$1600 & A40 & 302ms & 25100M & 26.88 & 16\\
        & TPVFormer\cite{huang2023tri} & ResNet-101 & 640$\times$1600  & A40 & 341ms & 29000M& 28.34 & 15\\
        & FlashOcc-M6\cite{yu2023flashocc} & SwinTransformer-B & 640$\times$1600  & RTX3090 & - & -& 45.50 & 7\\  
        & FlashOcc-M4 (TensorRT)\cite{yu2023flashocc} & ResNet-50 & 256$\times$704  & RTX3090 & 6.5ms & 2600M& 32.90 & -\\  
        & OccTransformer\cite{liu2024occtransformer} & InternImage-XL & - &V100 & - & - &49.23 & 6\\
        & FastOcc\cite{hou2024fastocc} & ResNet-101 & 640$\times$1600  &  V100 & 221ms & -& 40.75 & 11\\
        & FastOcc-Tiny\cite{hou2024fastocc} & ResNet-50 & 320$\times$1600  & V100 & 62.8ms & -& 34.21 & -\\
        \midrule
        \multirow{4}{*}{\makecell[l]{Coarse-to-fine}} 
        & PanoOcc\cite{wang2023panoocc} & ResNet-101 & 640$\times$1600  & A100 & 149ms & 35000M& 42.13 & 9\\
        & OctreeOcc\cite{lu2023octreeocc} & ResNet-101 & 900$\times$1600  & A40 & 386ms & 26500M& 44.02 & 8\\
        & SparseOcc (8f)\cite{liu2023fully} & ResNet-50 & 256$\times$704  & V100 & - & -& 30.90 & 13\\  
        & SparseOcc (1f)\cite{liu2023fully} & ResNet-50 & 256$\times$704  & V100 & 78.7ms & -& 27.00 & -\\  
		\bottomrule
	\end{tabular}
    }
    \label{table:deployment-friendly}
\end{table*}

The performance comparison of deployment-friendly methods on Occ3D-nuScenes dataset is shown in Table~\ref{table:deployment-friendly}. Due to the results being collected from different papers, with variations in backbone, image size, and computing platform, only some preliminary conclusions can be drawn. Generally, under similar experimental settings, coarse-to-fine methods outperform perspective decomposition methods in terms of performance due to fewer information losses, while perspective decomposition often exhibits better real-time performance and lower memory usage. Additionally, models that adopt heavier backbones and process larger images can achieve better accuracy but will also weaken the real-time performance. Although light versions of methods like FlashOcc~\cite{yu2023flashocc} and FastOcc~\cite{hou2024fastocc} get close to the requirements of the actual deployment, their accuracy needs to be further improved. For deployment-friendly methods, both the perspective decomposition strategy and the coarse-to-fine strategy are committed to continuously pursuing the reduction of computational load while maintaining the accuracy of 3D occupancy prediction.

\vspace{-0.03cm}
\section{Label-efficient methods}
\label{LE}

In the existing methods for creating precise occupancy labels \cite{wang2023openoccupancy, tian2023occ3d, tong2023scene}, there are two essential steps. The first one is collecting LiDAR point clouds that correspond to multi-view images and making semantic segmentation annotations. The other one is fusing multi-frame point clouds through complicated algorithms with the help of the tracking information of dynamic objects. Both steps are quite expensive, which limits the ability of occupancy networks to utilize the massive amounts of multi-view images from autonomous driving scenarios. Recently, Neural Radiance Field (Nerf)~\cite{mildenhall2021nerf} has been widely used in 2D image rendering. Several methods render the predicted 3D occupancy into 2D maps in Nerf-like ways and train the occupancy network without the participation of fine-grained annotation or LiDAR point clouds, which significantly reduces the cost of data annotation.

\subsection{Annotation-free methods}

\begin{figure}[!ht]
    \centering
    \includegraphics[width=\linewidth]{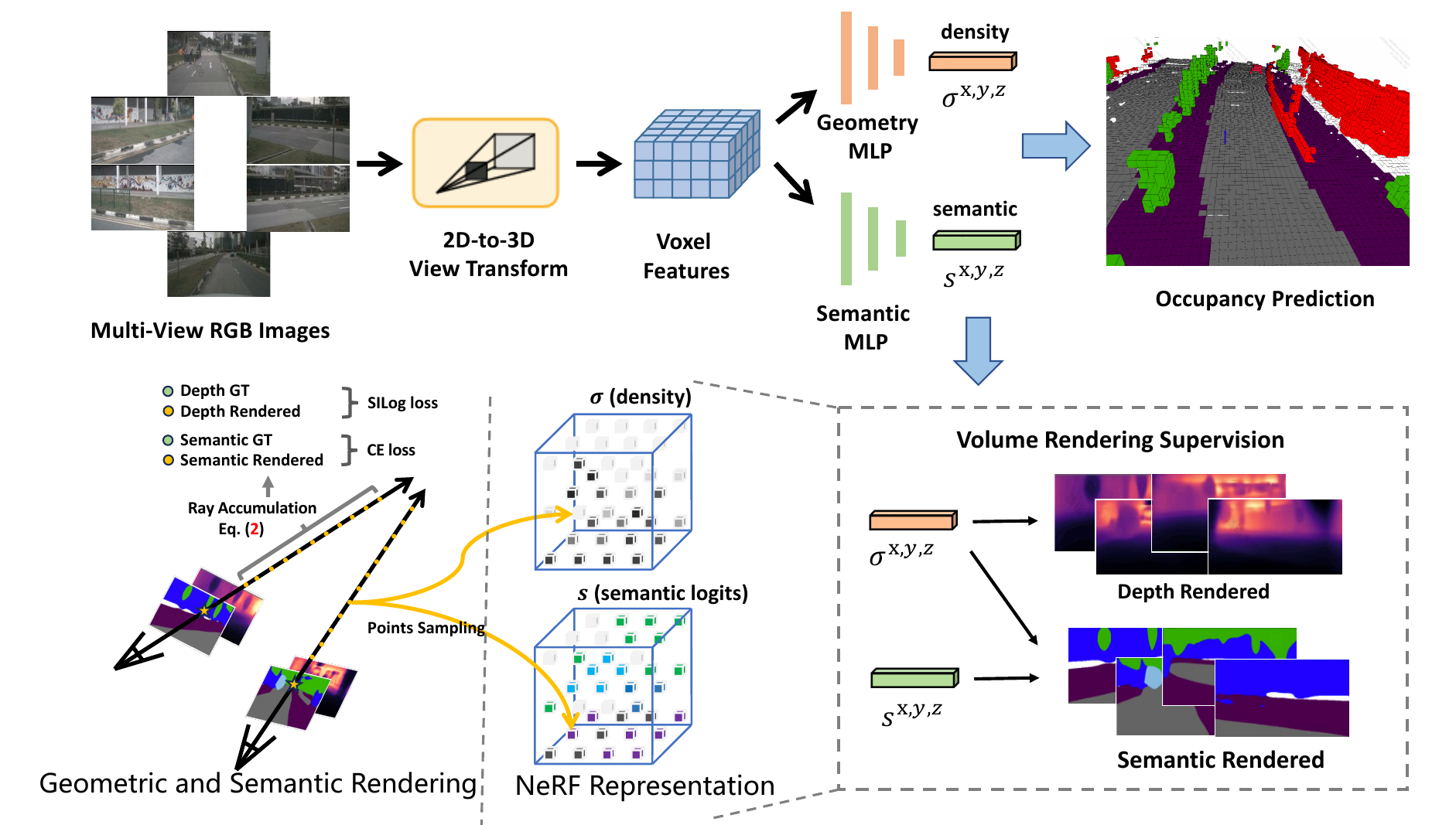}
    \caption{UniOcc~\cite{pan2023uniocc} supervises the rendered depth maps and semantic maps to learn the 3D occupancy.}
    \label{fig:uniocc}
\end{figure}

SimpleOccupancy~\cite{gan2023simple} first generates the explicit 3D voxel features of the scene from image features via view transform and then renders them into 2D depth maps following the Nerf-style way. The 2D depth maps are supervised by the sparse depth maps generated by LiDAR point clouds. The depth maps are also used to synthesize surround images for self-supervision.
UniOcc~\cite{pan2023uniocc} uses two separate MLPs to transform 3D voxel logits into the density of voxels and semantic logits of voxels. After that, UniOCC follows the general volume rendering to obtain the multi-view depth maps and semantic maps as shown in Figure~\ref{fig:uniocc}. Those 2D maps are supervised by the labels generated by segmented LiDAR point clouds. RenderOcc~\cite{pan2023renderocc} constructs a 3D volume representation similar to NeRF from multi-view images and employs advanced volume rendering techniques to generate 2D renderings, which can provide direct 3D supervision using only 2D semantic and depth labels. With this 2D rendering supervision, the model learns multi-view consistency by analyzing ray intersections from various camera frustums, thus gaining a deeper understanding of the geometric relationships in 3D space. Additionally, it introduces the concept of auxiliary rays to leverage rays from neighboring frames to enhance the multi-view consistency constraint of the current frame and develops a dynamic sampling training strategy to filter out misaligned rays. To address the imbalance between dynamic and static categories, OccFlowNet~\cite{boeder2024occflownet} futher introduces occupancy flow, predicting scene flow for each dynamic voxel based on 3D bounding boxes. With voxel flow, dynamic voxels can be moved to their correct positions in the temporal frames, eliminating the need for dynamic object filtering during rendering. During training, voxels that are correctly predicted and within the bounding boxes are transformed using flow to align with the target positions in the temporal frames, followed by grid alignment using distance-based weighted interpolation.

Due to the lack of geometric priors, vision-based 3D occupancy prediction faces significant challenges in achieving accurate predictions. RadOcc~\cite{zhang2024radocc} addresses this issue by exploring cross-modal knowledge distillation in this task, leveraging more powerful multi-modal models to guide the vision-based model during training. Specifically, RadOcc proposes a rendering-assisted distillation paradigm for 3D occupancy prediction, generating depth and semantic maps in the perspective view using differentiable volume rendering and introducing two novel consistency criteria between the rendering outputs of teacher and student models. The depth consistency loss aligns termination distributions of rendered rays, while the semantic consistency loss mimics intra-segment similarity guided by the visual foundation model. Vampire~\cite{xu2023regulating} explores the relationship between spatial occupancy and volume density in NeRF and proposes volumetric rendering as an intermediate 3D feature modulator in multi-camera setups. Vampire predicts occupancy at each position as volume density and accumulates intermediate 3D features onto the 2D plane for additional 2D supervision.

The above methods eliminate the need for explicit 3D occupancy annotations, substantially reducing the burden of manual annotations. However, they still rely on LiDAR point clouds to supply depth or semantic labels for supervising the rendered maps, which falls short of achieving a fully self-supervised framework for 3D occupancy prediction.

\begin{table*}[!ht]
    \caption{Performance of the Label-efficient methods on Occ3D-nuScenes dataset. The best one is highlighted in \textbf{bold}. $\dag$ denotes the annotation-free methods that only use rendered labels. ``Rank'' denotes the accuracy rank among all methods. Note that some experimental settings may not be exactly the same, and the results are just collected here for reference.}
    \renewcommand{\tabcolsep}{3pt}
    \centering
    \scalebox{0.85}{
    	\begin{tabular}{l|l|ccccccccccccccccc|c|c}
		\toprule
		Category & Method
        & \rotatebox{90}{\textcolor{nothers}{$\blacksquare$} others}
		& \rotatebox{90}{\textcolor{nbarrier}{$\blacksquare$} barrier}
		& \rotatebox{90}{\textcolor{nbicycle}{$\blacksquare$} bicycle}
		& \rotatebox{90}{\textcolor{nbus}{$\blacksquare$} bus}
		& \rotatebox{90}{\textcolor{ncar}{$\blacksquare$} car}
		& \rotatebox{90}{\textcolor{nconstruct}{$\blacksquare$} const. veh.}
		& \rotatebox{90}{\textcolor{nmotor}{$\blacksquare$} motorcycle}
		& \rotatebox{90}{\textcolor{npedestrian}{$\blacksquare$} pedestrian}
		& \rotatebox{90}{\textcolor{ntraffic}{$\blacksquare$} traffic cone}
		& \rotatebox{90}{\textcolor{ntrailer}{$\blacksquare$} trailer}
		& \rotatebox{90}{\textcolor{ntruck}{$\blacksquare$} truck}
		& \rotatebox{90}{\textcolor{ndriveable}{$\blacksquare$} drive. suf.}
		& \rotatebox{90}{\textcolor{nother}{$\blacksquare$} other flat}
		& \rotatebox{90}{\textcolor{nsidewalk}{$\blacksquare$} sidewalk}
		& \rotatebox{90}{\textcolor{nterrain}{$\blacksquare$} terrain}
		& \rotatebox{90}{\textcolor{nmanmade}{$\blacksquare$} manmade}
		& \rotatebox{90}{\textcolor{nvegetation}{$\blacksquare$} vegetation}
        & mIoU & Rank
		\\
		\midrule
        \multirow{6}{*}{Annotation-Free} & UniOcc\cite{pan2023uniocc} & \textbf{26.94} & \textbf{56.17} & \textbf{39.55} & \textbf{49.40} & \textbf{60.42} & \textbf{35.51} & \textbf{44.77} & 42.96 & 38.45 & \textbf{59.33} & \textbf{45.90} & \textbf{83.90} & 53.53 & \textbf{59.45} & \textbf{56.58} & \textbf{63.82} & \textbf{54.98} & \textbf{51.27} & 3\\
        & RenderOcc\cite{pan2023renderocc} & 4.84 & 31.72 & 10.72 & 27.67 & 26.45 & 13.87 & 18.20 & 17.67 & 17.84 & 21.19 & 23.25 & 63.20 & 36.42 & 46.21 & 44.26 & 19.58 & 20.72 & 26.11 & 17\\  
        & RenderOcc\dag\cite{pan2023renderocc} & 5.69 & 27.56 & 14.36 & 19.91 & 20.56 & 11.96 & 12.42 & 12.14 & 14.34 & 20.81 & 18.94 & 68.85 & 33.35 & 42.01 & 43.94 & 17.36 & 22.61 & 23.93 & -\\  
        & OccFlowNet\cite{boeder2024occflownet} & 8.00 & 37.60 & 26.00 & 42.10 & 42.50 & 21.60 & 29.20 & 22.30 & 25.70 & 29.70 & 34.40 & 64.90 & 37.20 & 44.30 & 43.20 & 34.30 & 32.50 & 33.86 & 12\\
        & OccFlowNet\dag\cite{boeder2024occflownet} & 1.60 & 27.50 & 26.00 & 34.00 & 32.00 & 20.40 & 25.90 & 18.60 & 20.20 & 26.00 & 28.70 & 62.00 & 27.20 & 37.80 & 39.50 & 29.00 & 26.80 & 28.42 & -\\
        & RadOcc\cite{zhang2024radocc} & 21.13 & 55.17 & 39.31 & 48.99 & 59.92 & 33.99 & 46.31 & \textbf{43.26} & \textbf{39.29} & 52.88 & 44.85 & 83.72 & \textbf{53.93} & 59.17 & 55.62 & 60.53 & 51.55 & 49.98 & 4\\
        \midrule
        \multirow{2}{*}{LiDAR-Free} & OccNeRF\cite{zhang2023occnerf} &-- & 0.83 & 0.82 & 5.13 & 12.49 & 3.50 & 0.23 & 3.10 & 1.84 & 0.52 & 3.90 & 52.62	&-- & 20.81 & 24.75 & 18.45 & 13.19 & 10.81 & 20\\
        & SelfOcc\cite{huang2023selfocc} & 0.00 & 0.15 & 0.66 & 5.46 & 12.54 & 0.00 & 0.80 & 2.10 & 0.00 & 0.00 & 8.25 & 55.49 & 0.00 & 26.30 & 26.54 & 14.22 & 5.60 & 9.30 & 21\\  
		\bottomrule
	\end{tabular}
    }
    \label{table:label-efficient}
\end{table*}

\subsection{LiDAR-free methods}

\begin{figure}[!ht]
    \centering
    \includegraphics[width=\linewidth]{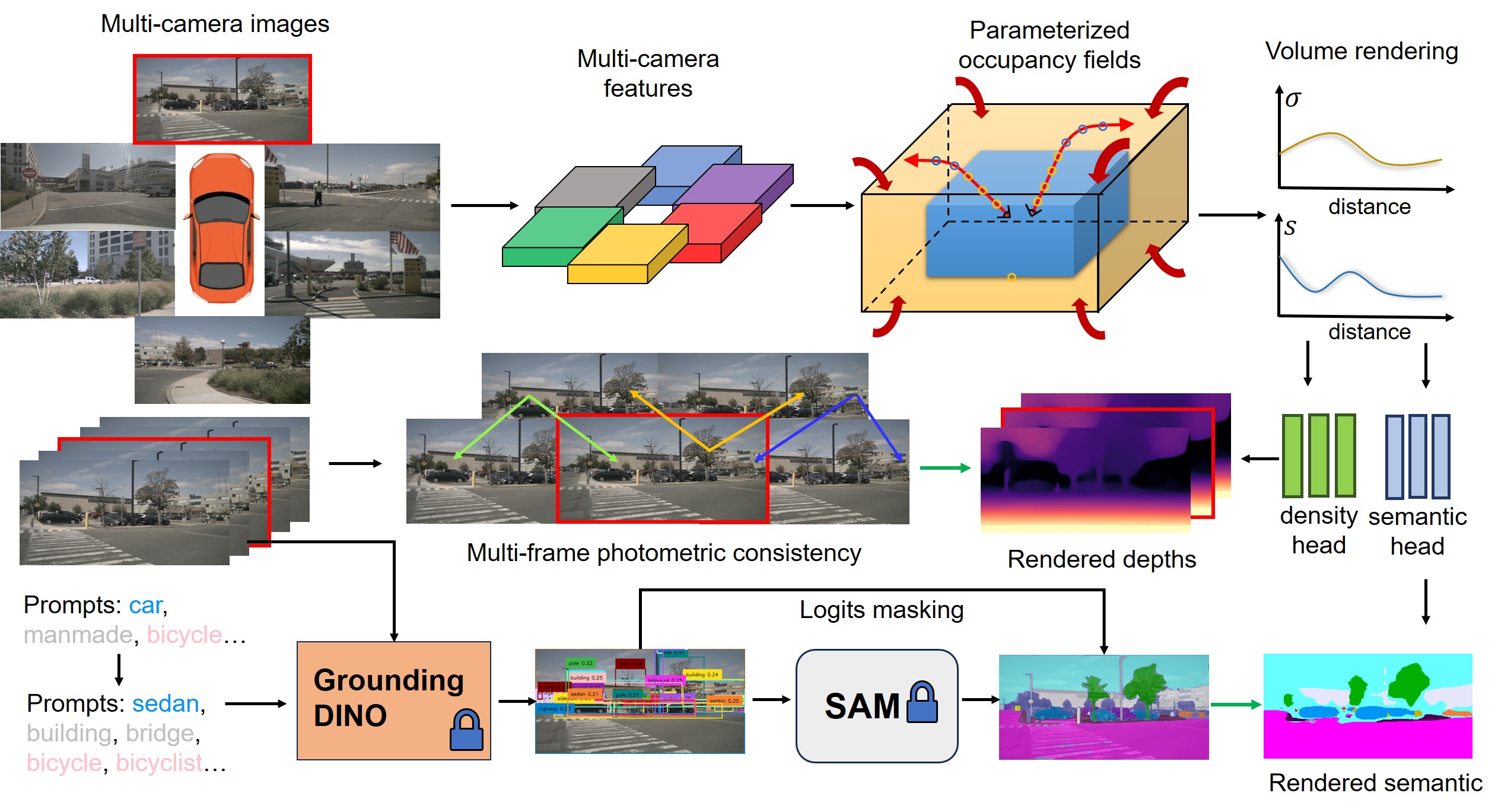}
    \caption{OccNerf~\cite{zhang2023occnerf} utilizes the photometric consistency and pretrained model to provide labels for the rendered maps.}
    \label{fig:occnerf}
\end{figure}

OccNerf~\cite{zhang2023occnerf} does not utilize the LiDAR point cloud for providing depth and semantic labels. Instead, as shown in Figure~\ref{fig:occnerf}, it addresses unbounded outdoor scenes using Parameterized Occupancy Fields, reorganizes the sampling strategy, and employs Volume Rendering to convert the occupancy field into multi-camera depth maps, ultimately supervised through multi-frame photometric consistency. Additionally, the method utilizes a pre-trained open vocabulary semantic segmentation model to generate 2D semantic labels, supervising the model to impart semantic information to the occupancy field. Behind the Scenes~\cite{wimbauer2023behind} uses a single view image sequence to reconstruct the driving scenes. It regards the frustum features of the input image as the density field and renders the synthesis of the other views. The whole model is trained by the specially designed image reconstruction loss. SelfOcc~\cite{huang2023selfocc} predicts the signed distance field values of the BEV or TPV features to render the 2D depth maps. Besides, the raw color and semantic maps are also rendered and supervised by the label generated from the sequence of multi-view images.

These approaches circumvent the necessity for depth or semantic labels from LiDAR point clouds. Instead, they leverage image data or pretrained models to obtain these labels, thereby achieving a genuine self-supervised framework for 3D occupancy prediction. Although these methods can achieve the training mode that best meets practical application expectations, further exploration is still needed to achieve satisfactory performance.

The performance comparison of label-efficient methods on Occ3D-nuScenes dataset is shown in Table~\ref{table:label-efficient}. Most of the Annotation-free methods use 2D rendering supervision as a complement to explicit 3D occupancy supervision and certain performance improvement is obtained. Among them, UniOcc~\cite{pan2023uniocc} and RadOcc~\cite{zhang2024radocc} even achieve excellent rankings of 3 and 4 respectively among all methods, fully demonstrating the annotation-free mechanism can facilitate the extraction of additional valuable information. When only 2D rendering supervision is employed, they can still achieve comparable accuracy, illustrating the feasibility of saving the cost of explicit 3D occupancy annotation. The LiDAR-free methods establish a comprehensive self-supervision framework for 3D occupancy prediction, further eliminating the need for labels and LiDAR data. However, due to the lack of precise depth and geometric information inherent in point clouds, their performance is greatly limited. 
 
\section{Future outlooks}
\label{FO}
%zjq
Driven by the approaches outlined above, we conclude present trends and propose several vital research directions with the potential to significantly advance the field of vision-based 3D occupancy prediction for autonomous driving from the perspectives of data, methodology and task.

\subsection{Data level}
Acquiring ample realistic driving data is paramount for enhancing the overall capability of autonomous driving perception systems. Data generation is a promising avenue, as it incurs no acquisition costs and offers the flexibility to manipulate data diversity as needed. While some methods~\cite{wen2023panacea,hu2023gaia} leverage prompts like text to control the content of the generated driving data, they can not guarantee the accuracy of the spatial information. In contrast, 3D Occupancy provides a fine-grained and actionable representation of scenes, facilitating controllable data generation and spatial information display when compared to point clouds, multi-view images, and BEV layout. WoVoGen~\cite{lu2023wovogen} proposes volume-aware diffusion that can map 3D occupancy to realistic multi-view images. After making modifications to 3D occupancy, such as adding a tree or replacing a car, the diffusion model will synthesize the corresponding new driving scene. The modified 3D occupancy records the 3D position information, which ensures the authenticity of the synthetic data. 

\begin{figure}[!ht]
    \centering
    \includegraphics[width=\linewidth]{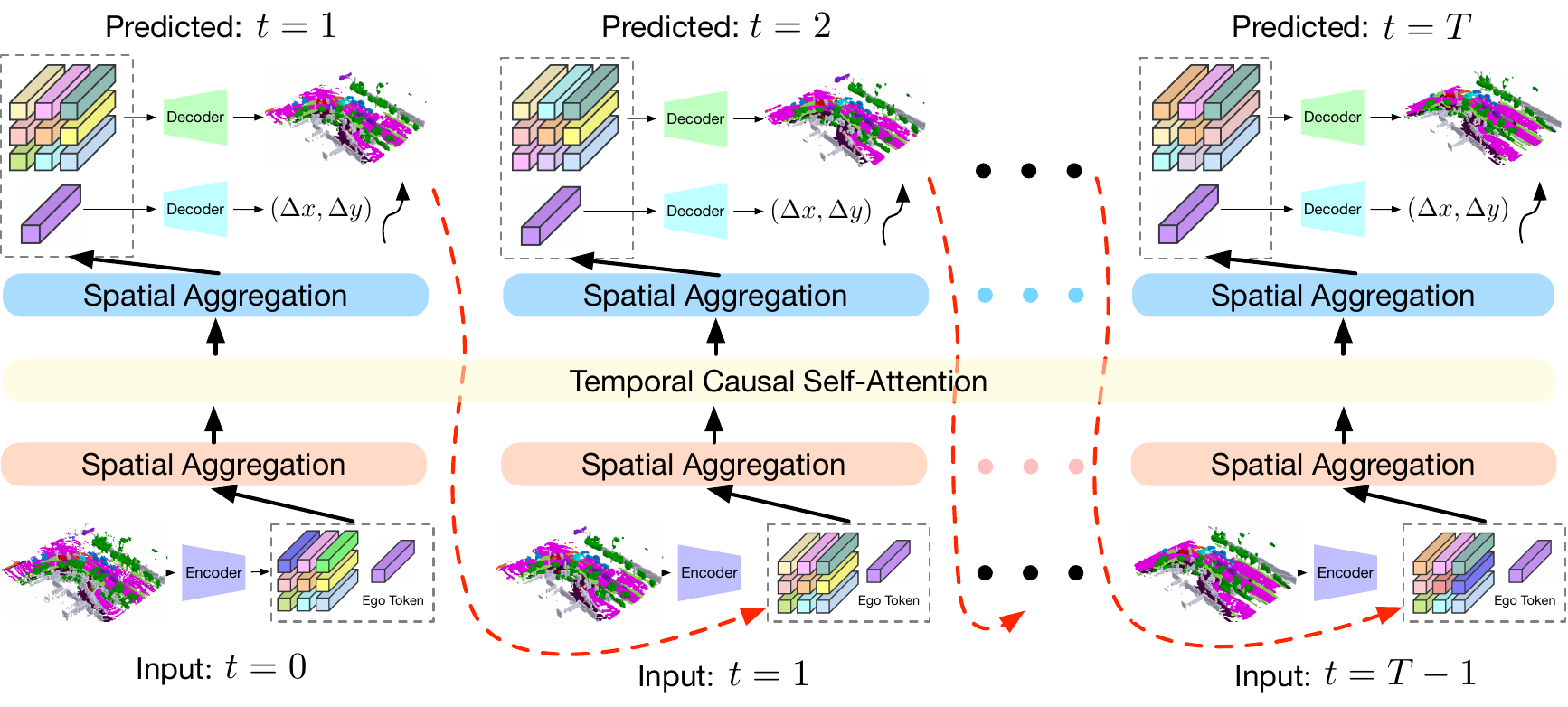}
    \caption{OccWorld~\cite{zheng2023occworld} chooses 3D occupancy data as the observation and prediction in the world model.}
    \label{fig:OccWorld}
\end{figure}

The world model for autonomous driving is increasingly prominent, offering a straightforward and elegant framework that enhances the model's capacity to comprehend the entire scene based on input observations of the environment and directly output suitable dynamic scene evolution data. Given its proficiency in representing the entire driving scene data in detail, leveraging 3D occupancy as the environmental observation in world model confers distinct advantages. As shown in Figure~\ref{fig:OccWorld}, OccWorld~\cite{zheng2023occworld} chooses 3D occupancy as the input of the world model and uses the GPT-like module to predict what the 3D occupancy data should be like in the future. UniWorld~\cite{min2023uniworld} utilizes the off-the-shelf BEV-based 3D occupancy model, but predicts the future 3D occupancy data by processing the past multi-view images, which also constructs a world model. However, regardless of the mechanism, there inevitably exists a domain gap between generated data and real data. To address this issue, one feasible approach is to integrate 3D occupancy prediction with the emerging 3D Artificial Intelligence Generated Content (3D AIGC) methods to generate more realistic scene data, while another approach is to combine domain adaptation methods to narrow down the domain gap.

\subsection{Methodology level}
When it comes to 3D occupancy prediction methods, there are ongoing challenges that require further attention across the categories we previously outlined: feature enhancement methods, deployment-friendly methods and label-efficient methods. Feature enhancement methods need to develop in a direction that significantly improving the performance while maintaining controllable computational resource consumption. Deployment-friendly methods should keep in mind that decreasing memory usage and latency while ensuring minimal performance degradation. Label-efficient methods should evolve towards reducing expensive annotation requirements while achieving satisfactory performance. The ultimate goal may be to achieve a unified framework that combines feature enhancement, deployment-friendliness, and label efficiency to meet the expectations of practical autonomous driving applications.

\begin{figure}[!ht]
    \centering
    \includegraphics[width=\linewidth]{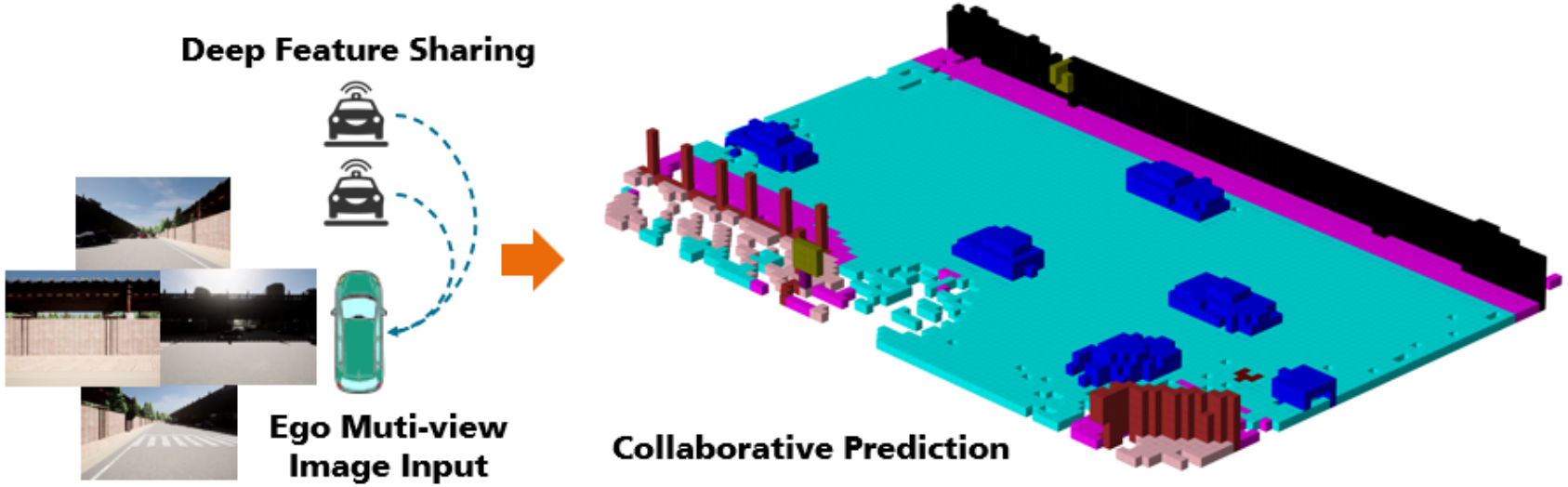}
    \caption{Multi-agent collaborative 3D occupancy prediction~\cite{song2024collaborative}.}
    \label{fig:v2x}
\end{figure}

Furthermore, the existing single-agent autonomous driving perception systems are inherently unable to address key issues such as sensitivity to occlusions, insufficient long-range perception capabilities, and limited field of view, making it challenging to achieve comprehensive environmental awareness. In order to overcome the bottleneck of single-agent, multi-agent collaborative perception method opens up a new dimension, allowing vehicles to share complementary information with other traffic elements to obtain a holistic perception of the surrounding environment. As shown in Figure~\ref{fig:v2x}, multi-agent collaborative 3D occupancy prediction method leverages the power of collaborative perception and learning for 3D occupancy prediction, enabling a deeper understanding of the 3D road environment by sharing features among connected automated vehicles. CoHFF~\cite{song2024collaborative} is the first vision-based framework for collaborative semantic occupancy prediction, which improves local 3D semantic occupancy predictions by hybrid fusion of semantic and occupancy task features, and compressed orthogonal attention features shared between vehicles, significantly outperforming single-vehicle systems in performance. However, such methods often require communication with multiple agents simultaneously, facing a contradiction between accuracy and bandwidth. Therefore, identifying which agents require coordination the most, as well as determining the most valuable collaborative areas to achieve the optimal balance between accuracy and speed, is an interesting research direction.

\subsection{Task level} 
In current 3D occupancy benchmarks, certain categories possess explicit semantics, such as ``car'', ``pedestrian'' and ``truck''. Conversely, the semantics of other categories like ``manmade'' and ``vegetation'' tend to be vague and general. These categories encompass a wide range of undefined semantics and ought to be subdivided into more fine-grained categories to provide detailed descriptions of driving scenes. Additionally, for unknown categories that have not been seen before, they are generally treated as general obstacles, without the ability to flexibly expand new category perception based on human prompts. The open-vocabulary task has demonstrated strong performance in 2D image perception for this issue and can be extended to improve the 3D occupancy prediction task. OVO~\cite{tan2023ovo} proposes a framework that supports the open-vocabulary 3D occupancy prediction. It utilizes a frozen 2D segmenter and text encoder to obtain the open-vocabulary semantic reference. Afterward, alignments at three different levels are employed to distill the 3D occupancy model, enabling it to make open-vocabulary predictions. POP-3D~\cite{vobecky2024pop} designs a self-supervised framework incorporating three modalities aided by a strong pretrained vision-language model. It facilitates open-vocabulary tasks such as zero-shot occupancy segmentation and text-based 3D retrieval.

Perceiving dynamic changes in the surrounding environment is crucial for the safe and reliable execution of downstream tasks in autonomous driving. While 3D occupancy prediction can provide dense occupancy representations of large-scale scenes based on current observations, they mostly confine to representing the current 3D space and do not consider the future states of surrounding objects along the temporal axis. Recently, several methods have been proposed to further consider the temporal information and introduce the 4D occupancy prediction task, which is more practical in real autonomous driving scenarios. Cam4DOcc~\cite{ma2023cam4docc} for the first time establishes a novel benchmark for 4D occupancy prediction using the widely utilized nuScenes dataset~\cite{caesar2020nuscenes}. This benchmark includes diverse metrics tailored to assess the occupancy prediction of general movable objects (GMO) and general static objects (GSO) separately. Moreover, it offers several baseline models to illustrate the construction of a 4D occupancy prediction framework. Although the open-vocabulary 3D occupancy prediction task and the 4D occupancy prediction task aim to enhance the perception capability of autonomous driving in open dynamic environments from different views, they are still treated as independent tasks to be optimized. The modularized task-based paradigm, where multiple modules have inconsistent optimization objectives, can lead to information loss and cumulative errors. Integrating open-set dynamic occupancy prediction with end-to-end autonomous driving task to directly map raw sensor data to control signals is a promising research direction.

\begin{acknowledgement}
This work is partly supported by the National Natural Science Foundation of China (62022011), the Research Program of State Key Laboratory of Software Development Environment, and the Fundamental Research Funds for the Central Universities.
\end{acknowledgement}

\begin{competinginterest}
The authors declare that they have no competing interests or financial conflicts to disclose.
\end{competinginterest}

\bibliographystyle{fcs}
\bibliography{ref}

\end{document}